\documentclass[10pt,journal,twoside]{tex/IEEEtran}
\usepackage{epsfig}
\usepackage{graphicx}
\usepackage{enumitem}
\usepackage{sidecap}
\usepackage{framed}
\usepackage{tabularx}

\graphicspath{{figures/}}

\ifCLASSINFOpdf
\else
\fi

\usepackage[cmex10]{amsmath}
\usepackage{amssymb}
\usepackage{mathtools}
\usepackage{url}
\renewcommand{\abovecaptionskip}{0.5cm}

\begin{document}
%\linenumbers
%
% paper title
% can use linebreaks \\ within to get better formatting as desired
% Do not put math or special symbols in the title.
\title{Visual object tracking performance measures revisited}
%
%
% author names and IEEE memberships
% note positions of commas and nonbreaking spaces ( ~ ) LaTeX will not break
% a structure at a ~ so this keeps an author's name from being broken across
% two lines.
% use \thanks{} to gain access to the first footnote area
% a separate \thanks must be used for each paragraph as LaTeX2e's \thanks
% was not built to handle multiple paragraphs
%

\author{Luka~\v{C}ehovin~\IEEEmembership{Member,~IEEE,}
        Ale\v{s}~Leonardis,~\IEEEmembership{Member,~IEEE,}
        and~Matej~Kristan,~\IEEEmembership{Member,~IEEE}% <-this % stops a space
\IEEEcompsocitemizethanks{\IEEEcompsocthanksitem L.~\v{C}ehovin and M.~Kristan are with the Faculty of
Computer and Information Science at the University of Ljubljana, Ljubljana, Slovenia. E-mail: \{luka.cehovin,matej.kristan\}@fri.uni-lj.si. \protect\\
\IEEEcompsocthanksitem A.~Leonardis is with the School of Computer Science and Centre for Computational Neuroscience and Cognitive Robotics, University of Birmingham, United Kingdom and with the Faculty of Computer and Information Science, University of Ljubljana, Slovenia. E-mail: a.leonardis@cs.bham.ac.uk, ales.leonardis@fri.uni-lj.si\protect\\
}% <-this % stops a space
\thanks{}}

% note the % following the last \IEEEmembership and also \thanks - 
% these prevent an unwanted space from occurring between the last author name
% and the end of the author line. i.e., if you had this:
% 
% \author{....lastname \thanks{...} \thanks{...} }
%                     ^------------^------------^----Do not want these spaces!
%
% a space would be appended to the last name and could cause every name on that
% line to be shifted left slightly. This is one of those "LaTeX things". For
% instance, "\textbf{A} \textbf{B}" will typeset as "A B" not "AB". To get
% "AB" then you have to do: "\textbf{A}\textbf{B}"
% \thanks is no different in this regard, so shield the last } of each \thanks
% that ends a line with a % and do not let a space in before the next \thanks.
% Spaces after \IEEEmembership other than the last one are OK (and needed) as
% you are supposed to have spaces between the names. For what it is worth,
% this is a minor point as most people would not even notice if the said evil
% space somehow managed to creep in.

% The paper headers
\markboth{IEEE Transactions on Image Processing}%
{\v{C}ehovin \MakeLowercase{\textit{et al.}}: Visual object tracking performance measures revisited}
% The only time the second header will appear is for the odd numbered pages
% after the title page when using the twoside option.
% 
% *** Note that you probably will NOT want to include the author's ***
% *** name in the headers of peer review papers.                   ***
% You can use \ifCLASSOPTIONpeerreview for conditional compilation here if
% you desire.

% If you want to put a publisher's ID mark on the page you can do it like
% this:
%\IEEEpubid{0000--0000/00\$00.00~\copyright~2012 IEEE}
% Remember, if you use this you must call \IEEEpubidadjcol in the second
% column for its text to clear the IEEEpubid mark.

% make the title area
\maketitle

% As a general rule, do not put math, special symbols or citations
% in the abstract or keywords.
\begin{abstract}
The problem of visual tracking evaluation is sporting a large variety of performance measures, and largely suffers from lack of consensus about which measures should be used in experiments. This makes the cross-paper tracker comparison difficult. Furthermore, as some measures may be less effective than others, the tracking results may be skewed or biased towards particular tracking aspects. In this paper we revisit the popular performance measures and tracker performance visualizations and analyze them theoretically and experimentally. We show that several measures are equivalent from the point of information they provide for tracker comparison and, crucially, that some are more brittle than the others. Based on our analysis we narrow down the set of potential measures to only two complementary ones, describing accuracy and robustness, thus pushing towards homogenization of the tracker evaluation methodology. These two measures can be intuitively interpreted and visualized and have been employed by the recent Visual Object Tracking (VOT) challenges as the foundation for the evaluation methodology.
\end{abstract}

% Note that keywords are not normally used for peerreview papers.

% For peer review papers, you can put extra information on the cover
% page as needed:
% \ifCLASSOPTIONpeerreview
% \begin{center} \bfseries EDICS Category: 3-BBND \end{center}
% \fi
%
% For peerreview papers, this IEEEtran command inserts a page break and
% creates the second title. It will be ignored for other modes.
\IEEEpeerreviewmaketitle

\section{Introduction}

\IEEEPARstart{V}{isual} tracking is one of the rapidly evolving fields of computer vision. Every year, literally dozens of new tracking algorithms are presented and evaluated in journals and at conferences. When considering the evaluation of these new trackers and comparison to the state-of-the-art, several questions arise. Is there a standard set of sequences that we can use for the evaluation? Is there a standardized evaluation protocol? What kind of performance measures should we use? Unfortunately, there are currently no definite answers to these questions. Unlike some other fields of computer vision, like object detection and classification \cite{Everingham2009}, optical-flow computation \cite{Baker2010} and automatic segmentation \cite{Alpert2007}, where widely adopted evaluation protocols are used, visual tracking is still largely lacking these properties.

The absence of homogenization of the evaluation protocols makes it difficult to rigorously compare trackers across publications and stands in the way of faster development of the field. The authors of new trackers typically compare their work against a limited set of related algorithms due to the difficulty of adapting these for their own use in the experiments. One of the issues here is the choice of tracker performance evaluation measures, which seems to be almost arbitrary in the tracking literature. Worse yet, an abundance of these measures are currently in use~\cite{Wang2011,Wu2013,Nawaz2012,Smeulders2013}. Because of this, experiments in many cases offer a limited insight into the tracker's performance, and prohibit comparisons across different papers.

In contrast to the existing works on evaluation of single-target visual trackers, that focus on benchmarking visual trackers without considering the selection of good measures, or propose new complex measures, we take a different approach. We investigate various popular performance evaluation measures using theoretically provable relations between them as well as systematic experimental analysis. We discuss their pitfalls and show that, from a standpoint of tracker comparison, many of the widely used measures are in fact equivalent. In addition we prove a direct relation of two complex recently proposed performance measures with the basic performance measures thus allowing their analysis in terms of the basic performance measures. Since several measures reflect the same aspects of tracking performance, combining those provides no additional performance insights and in fact introduces bias towards a particular aspect of performance to the result. We identify complementary measures that are sensitive to two different aspects of trackers performance and demonstrate their practical interpretation on a large-scale experiment. We emphasize that the goal of our analysis is therefore not to rank state-of-the-art tracking algorithms and make claims on which is better, but to homogenize the tracking performance evaluation methodology and increase the interpretability of results.

%The main goal of our work is to investigate existing measures for short-term single-object visual tracking and propose a good subset of measures (in our final case a pair) that can be used for tracker description and comparison. We base our investigation on an experimental analysis using a large sequence dataset and numerous diverse trackers and observe the behavior of individual measures in a systematic way, therefore we argue that the systematic principle is addressed. Besides that we also employ the theoretical principle where possible with two derivations of equality of measures that are included in our analysis. Note that the theoretical analysis is mainly limited by a loosely defined groundtruth guidelines in the community, however, this issue goes beyond the scope of our paper.

In our work we focus on the problem of performance evaluation in monocular single-target visual tracking that does not contain complete disappearance of the target from the scene that would require later re-detection; this kind of tracking scenario is also known as {\em short-term} single-target visual tracking in contrast to single-target long-term tracking (target has to be re-detected)~\cite{Kalal2010,Ma2015,Hong2015} and multi-target tracking (multiple targets)~\cite{Kasturi2009,Carvalho2012}. It is worth noting that our findings have been so far already used as the foundation of the evaluation methodology of the recent Visual Object Tracking challenges VOT2013~\cite{Kristan2013} as well as VOT2014~\cite{Kristan2014}.

\subsection{Related work}
\label{related}

Until recently the majority of papers that address performance evaluation in visual tracking were concerned with multi-target tracking scenarios~\cite{Smith2005,Kasturi2009,Black2003,Brown2005,Kao2009,Manohar2006,Bashir2006,Carvalho2012,Leichter2013}. Single-target tracking is, at least theoretically, a special case of multi-target tracking, however, because of the nature of the target domain, there is a crucial difference in the focus of the evaluation. In multi-target tracking, the focus is on the correctness of target identity assignments for a varying number of targets as well as the accuracy of these detections. The algorithms are often focused on a particular tracking domain, which is typically people or vehicle tracking for surveillance~\cite{Black2003,Brown2005,Jaynes2002}, animal groups tracking~\cite{Khan2005} or sports tracking~\cite{Kristan2009}, to name a few, which means that tracking in multi-object scenarios involves a lot of domain-specific prior knowledge. A well known PETS workshop (e.g.~\cite{Bashir2006}) has also been organized yearly for more than a decade with the main focus on performance evaluation of surveillance and activity recognition algorithms.

On the other hand, single-target visual tracking evaluation focuses on the accuracy of the tracker, as well as its robustness and generality. The goal is to demonstrate the tracking performance on a wide range of challenging scenarios (various types of objects, lighting conditions, camera motions, signal noise, etc.). In this respect, Wang et al.~\cite{Wang2011} compared several trackers using center error and overlap measures. Their research is focused primarily on investigating strengths and weaknesses of a limited set of trackers. In~\cite{Wu2013} authors perform an experimental comparison of several trackers. The performance measures in this case are chosen without theoretical justification which results in a poor qualitative analysis of the results. 
Nawaz and Cavallaro~\cite{Nawaz2012} have presented a system for evaluation of visual trackers that aims at addressing the real-world conditions in sequences. The system can simulate several real-world sources of noisy input, such as initialization noise, image noise and changes in the frame-rate. They have also proposed a new performance measure to address the trackers scoring, but the measure was introduced  without theoretical analysis of its properties. While such a measure may look like a good tool for ranking trackers, it cannot answer a simple question of in which aspect one tracker was better than the other. 
These recent experimental evaluations show the need for a better evaluation of visual trackers, however, none of them seems to address an important prerequisite for such evaluation, that is, what subset of the many available measures should be used for the evaluation. Frequently, multiple measures are used to cover multiple aspects of tracking performance without considering the fact that some measures describe the same aspects which leads to bias of the results. Instead, the selection should be grounded in a prior analysis of performance measures which is the main focus of this paper. Recently, Smeulders et al.~\cite{Smeulders2013} provided an experimental survey of several recent trackers together with an analysis of several performance measures. Their methodology and the general disposition in this aspect are similar to ours in terms that they search for multiple measures that describe different aspects of tracking performance. However, even though they do not explicitly acknowledge the fact that they address long-term tracking, their selection of measures and the dataset is from the start biased in favor of detection-based tracking algorithms, which also affects their results and derived conclusions.

Finally, evaluation of tracker performance without ground-truth annotations has been investigated by Wu et al.~\cite{Wu2010}, where the authors propose to use time-reversible nature of physical motion. As noted by SanMiguel et al.~\cite{SanMiguel2012}, this approach is not suitable for longer sequences. They propose to extend the approach using failure detection based on the uncertainty of the tracker. The problem is that the method has to be adapted to each tracker specifically and is useful only for investigative, but not for comparative purposes. An interesting approach to tracker comparison has also been recently proposed by Pang and Habin~\cite{Pang2013}. They aggregate existing experiments, published in various articles, in a page-rank fashion to form a ranking of trackers. The authors acknowledge that their meta-analysis approach is not appropriate for ranking recently published trackers. Furthermore, their approach does not remove bias that comes from correlation of multiple performance measures, which is one of the goals of our work. 

\subsection{Our approach and contributions}

In this paper we do not intend to propose new performance measures. Rather than doing this, we focus on narrowing the wide variety of existing measures for single-target tracking performance evaluation to only a few complementary ones. We claim a four-fold contribution: (1) We provide a detailed survey and experimental analysis of popular performance measures used in single-target tracking evaluation. (2) We show by experimental analysis that there exist clusters of performance measures that essentially indicate the same aspect of trackers performance. (3) By considering the theoretical aspects of existing measures as well as the experimental analysis we identify a subset of the two most suitable (complementary) measures that characterize trackers performance within the accuracy and robustness context as well as a simple and intuitive visualization of the selected pair of measures, and (4) we introduce the concept of theoretical tracker and propose four such trackers as guides in interpretation of tracker performance.

Our experimental analysis has been carried out in a form of a large-scale comparative experiment with $16$ state-of-the-art trackers and $25$ video sequences of common visual tracking scenarios. While the primary goal of this paper is not benchmarking trackers, we provide the performance results of the tested trackers on the two selected performance measures as a sideproduct of our experiment. We also intend to provide detailed results of the experiment (groundtruth and raw trajectories) as a side-product of our research\footnote{Raw data is available at \url{http://go.vicos.si/performancemeasures}.} for further study by other researchers. 

Preliminary results reported in this paper have been published in our conference paper~\cite{Cehovin2014}. This paper extends~\cite{Cehovin2014} in several ways. The related work has been significantly extended with the recent work in performance evaluation. The theoretical survey has been extended and proofs of reformulation of complex performance measures (e.g. CoTPS~\cite{Nawaz2012} and AUC~\cite{Wu2013,Nawaz2012}) in terms of the basic measures have been added. A new fragmentation indicator has been proposed to complement the analysis of failure rate measure. The experimental analysis has been extended by adding three state-of-the-art trackers. Two new theoretical trackers have been proposed to aid analysis of the selected performance measures. Guidelines have been set on automatic interpretation of sequence properties from the results of theoretical trackers.

The rest of the paper is organized as follows: Section \ref{measures} gives an overview of the current state of short-term single-target tracking performance evaluation measures. We describe our experimental setup and discuss the findings of the experiment in Section \ref{result}, where we also propose our selection of good measures together with several insights. Finally, we draw concluding remarks in Section \ref{conclusion}.

\section{Performance measures}
\label{measures}

\begin{figure}
    \centering
    \includegraphics[width = 0.9\columnwidth]{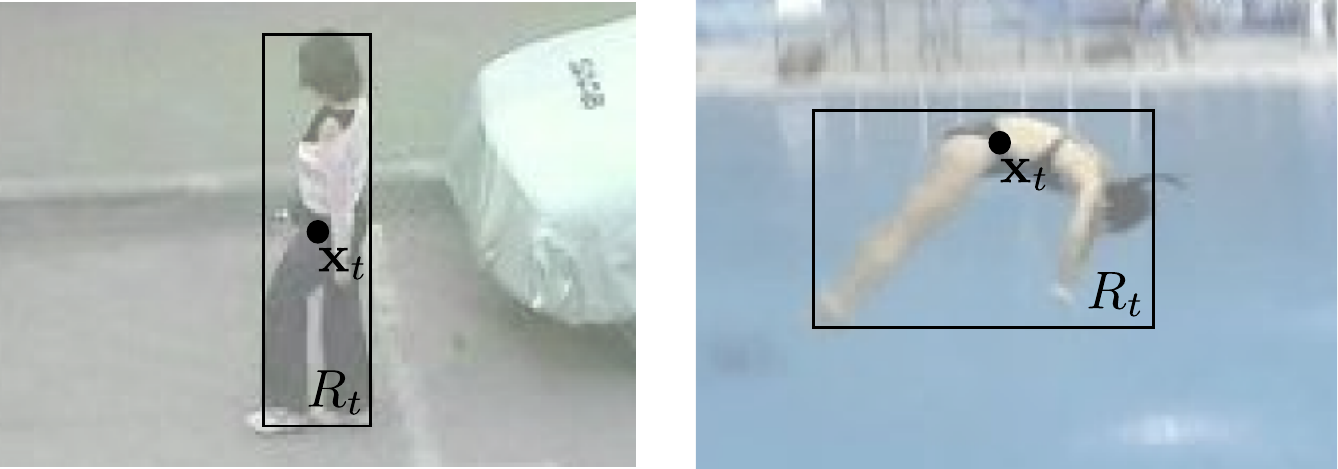}
    \caption{Two examples of an annotation for a single frame from the {\em woman} and the {\em driver} sequence. In the left example the center of the object can be estimated using the centroid of $R_t$, which is not true in the second case.}
    \label{fig:annotations}
\end{figure}

There are several performance measures that have become popular in single-target visual tracking evaluation and are widely used in the literature, however, none of them is a {\em de-facto} standard. As all of these measures assume that manual annotations are given for a sequence, we first establish a general definition of an object state description in a sequence with length $N$ as:

\begin{equation}
\label{eq:annotation}
\Lambda = \{ (R_t, \mathbf{x}_t) \}_{t = 1}^{N}, 
\end{equation}

\noindent where $\mathbf{x}_t \in \mathcal{R}^2$ denotes a center of the object and $R_t$ denotes the region of the object at time $t$. In practice the region is usually described by a bounding box (that is most commonly axis-aligned), however, a more complex shape could be used for a more accurate description. An example of two single-frame annotations can be seen in Figure \ref{fig:annotations}. In some cases the annotated center can be automatically derived from the region, but for some articulated objects, the centroid of region $R_t$ does not correspond to $\mathbf{x}_t$, therefore it is best to separately annotate $\mathbf{x}_t$. 

Performance measures aim at summarizing the extent to which the tracker's predicted annotation $\Lambda_T$ agrees with the ground truth annotation, i.e., $\Lambda_G$. 

\subsection{Center error}
\label{distance}

Perhaps the oldest means of measuring performance, which has its roots in aeronautics, is the center prediction error. This is still a popular measure~\cite{Ross2008,Babenko2011,Adam2006,Kwon2010,Wu2013,Wang2013,Yang2014a} and it measures the difference between the target's predicted center from the tracker and the ground-truth center.

\begin{equation}
\label{eq:distance}
\Delta(\Lambda^G, \Lambda^T) = \left\lbrace \delta_t \right\rbrace_{t = 1}^{N}, \ \ \delta_t = \| \mathbf{x}_t^G -\mathbf{x}_t^T \|.
\end{equation}

The popularity of center prediction measure comes from its minimal annotation effort, i.e., only a single point per frame. The results are usually shown in a plot, as in Figure~\ref{fig:overlap_vs_error} or summarized as average error (\ref{eq:avgdistance}), or root-mean-square-error (\ref{eq:rmse}):

\begin{equation}
\label{eq:avgdistance}
\Delta_\mu(\Lambda^G, \Lambda^T) = \frac{1}{N} \sum_{t=1}^N \delta_t,
\end{equation}

\begin{equation}
\label{eq:rmse}
\textrm{RMSE}(\Lambda^G, \Lambda^T) = \sqrt{\frac{1}{N} \sum_{t=1}^N \| \mathbf{x}_t^G -\mathbf{x}_t^T \|^2 }.
\end{equation}

One drawback of this measure is its sensitivity to subjective annotation (i.e., where exactly is the target's center). This sensitivity largely comes from the fact that the measure completely ignores the target's size and does not reflect the apparent tracking failure~\cite{Nawaz2012}. To remedy this, a normalized center error $\widehat{\Delta}(\cdot,\cdot)$ is used instead, e.g.~\cite{Bao2012,Smeulders2013}, in which the center error at each frame is divided by the tacker-predicted visual size of the target, $size(R_t^G)$,

\begin{equation}
\label{eq:normavgdistance}
\widehat{\Delta}(\Lambda^G, \Lambda^T) = \left\lbrace \widehat{\delta}_t \right\rbrace_{t = 1}^{N}, \ \ \widehat{\delta}_t = \| \frac{ \mathbf{x}_t^G -\mathbf{x}_t^T }{ size(R_t^G) } \|.
\end{equation}

Nevertheless, despite the normalization, the measure may give misleading results as the center error is reduced proportionally to the estimated target size. Furthermore, when the tracker fails and is drifting over a background, the actual distance between the annotated and reported center, combined with the estimated size (which can be arbitrarily large) overpowers the averaged score which does not properly reflect the important information that the tracker has failed.

\subsection{Region overlap}
\label{overlap}

The normalization problem is rather well addressed by the overlap-based measures~\cite{Zhang2012,Godec2011,Smeulders2013}. These measures require region annotations and are computed as an overlap between predicted target's region form the tracker and the ground-truth region:

\begin{equation}
\label{eq:overlap}
\Phi(\Lambda^G, \Lambda^T) = \left\lbrace \phi_t \right\rbrace _{t = 1}^{N} \ , \ \phi_t = \frac{R_t^G \cap R_t^T}{R_t^G \cup R_t^T}.
\end{equation}

\begin{figure}
    \centering
    \includegraphics[width = \columnwidth]{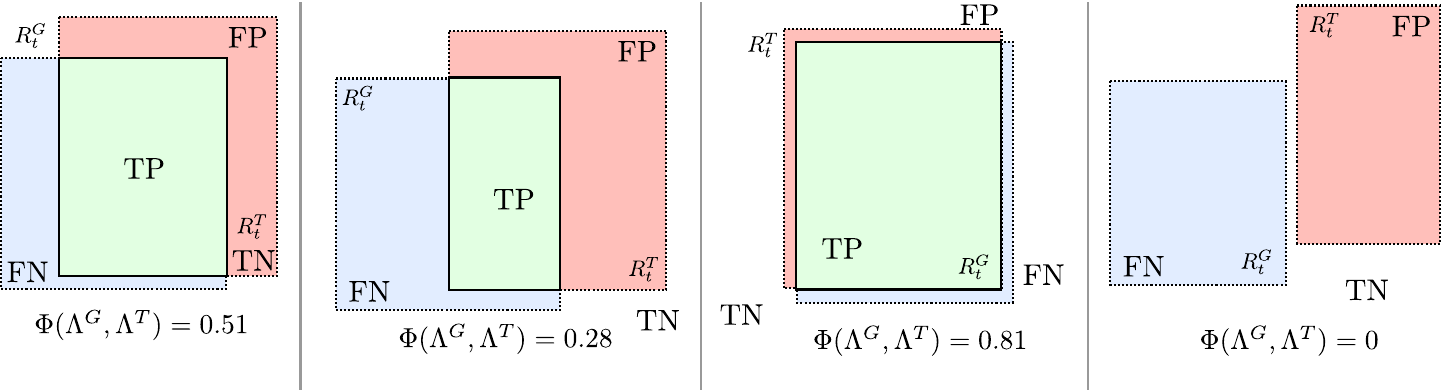}
    \caption{An illustration of the overlap of ground-truth region with the predicted region for four different configurations. }
    \label{fig:overlap}
\end{figure}

An appealing property of region overlap measures is that they account for both position and size of the predicted and ground-truth bounding boxes simultaneously, and do not result in arbitrary large errors at tracking failures, as is the case on center-based error measures. In fact, once the tracker drifts to the background, the measure becomes zero, regardless of how far from the target the tracker is currently located. In terms of pixel classification (see Figure~\ref{fig:overlap}), the overlap can be interpreted as

\begin{equation}
\label{eq:overlap2}
\frac{R_t^G \cap R_t^T}{R_t^G \cup R_t^T} = \frac{TP}{TP + FN + FP}, 
\end{equation}

\noindent a formulation similar to the F-measure in information retrieval, which can be written as $F = \frac{2 TP}{2 TP + FN + FP}$. Another closely related measure, used in tracking to account for un-annotated object occlusions is precision~\cite{Godec2011}, i.e. $\frac{TP}{TP + FP}$.

\begin{figure}[h]
    \centering
    \includegraphics[width = 0.9\columnwidth]{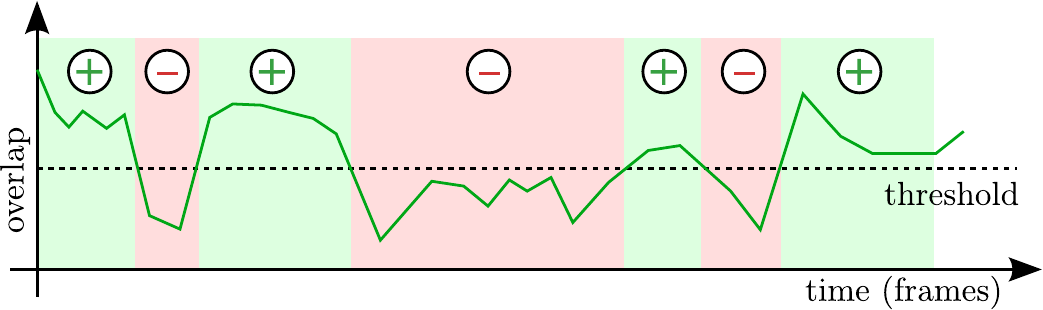}
    \caption{An illustration of overlap being used as a detection measure. The plus signs mark the intervals with positive detections (overlap above threshold), while minus signs mark the intervals with negative detections (interval below threshold).}
    \label{fig:detection}
\end{figure}

The overlap measure is summarized over an entire sequence by an {\em average overlap} (e.g. in~\cite{Wang2013,Wu2013}) that is defined as an average value of all region overlaps in the sequence

\begin{equation}
\label{eq:averageoverlap}
\bar{\phi} = \sum_t \frac{\phi_t}{N}.
\end{equation}

Another measure based on region overlap is number of correctly tracked frames $N_{\tau} = \sum_t \| \left\lbrace t | \phi_t > \tau \right\rbrace _{t = 1}^{N} \|$, where $\tau$ denotes a threshold on the overlap. This approach comes from the object detection community~\cite{Everingham2009}, where the overlap threshold for a correctly detected object is set to $\tau = 0.5$. The same threshold is often used for tracking performance evaluation, e.g. in~\cite{Zhang2012} and \cite{Wang2011}, however, this number is too high for general purpose tracking evaluation. As seen in Figure~\ref{fig:overlap} this threshold is reached even for visually well overlapping rectangles. This is especially problematic when considering non-rigid articulated objects. 

To make the final score more comparable across a set of sequences of different lengths, the number of correctly tracked frames is divided by the total number of frames
% TODO: citat

\begin{equation}
\label{eq:ppositive}
P_{\tau} (\Lambda^G, \Lambda^T) = \frac{ \| \left\lbrace t | \phi_t > \tau \right\rbrace _{t = 1}^{N} \| }{N}.
\end{equation}

The $P_{\tau}$, also known as {\em percentage of correctly tracked frames}, is a frame-wise definition of the {\em true-positive} score, an interpretation that has become popular in tracking evaluation with the advent of tracking-by-detection concept. As noted in~\cite{Smeulders2013}, the F-measure is another score that can be used in this context, however, it is worth noting that the detection based measures disregard the sequential nature of the tracking problem. As it is illustrated in Figure \ref{fig:detection}, these measures do not necessarily account for complete trajectory reconstruction which is an important aspect in many tracking applications.

The most popular measures for multi-target tracking performance, the Multiple Object Tracking Precision (MOTP) and Multiple Object Tracking Accuracy (MOTA)~\cite{Kasturi2009} can also be seen in the context of single-object short-term tracking as an extension of region overlap measures. MOTP measure is defined as average overlap over all objects on all frames, taking into account different number of objects that are visible at different frames, i.e. 

\begin{equation}
\label{eq:motp}
MOTP = \frac{\sum_{i = 1}^{M} \sum_{t = 1}^{N} \phi_{i, t}}{ \sum_{t = 1}^{N} M_{t} },
\end{equation}

\noindent where $M$ denotes the number of different objects in the entire sequence and $M_t$ denotes the number of visible objects at frame $t$. In single-target short-term tracking $M = M_t = 1$, therefore MOTP can be simplified to an average overlap measure, defined in equation (\ref{eq:averageoverlap}) earlier in this section. The MOTA measure, on the other hand, takes into account three components that account for accuracy of multiple-object tracking algorithm: number of misses, number of false alarms and number of identity switches, i. e.

\begin{equation}
\label{eq:mota}
MOTA = 1 - \frac{\sum_{t = 1}^{N} ( c_m MI_t + c_f FP_t + c_s SW_t )}{ \sum_{t = 1}^{N} N_{t}^G },
\end{equation}

\noindent where $MI_t$ denotes the number of misses, $FP_t$ denotes the number of wrong detections, $SW_t$ denotes the number of identity switches, $c_m$, $c_f$, and $c_s$, are weighting constants and $N_{t}^G$ denotes the number of annotated objects at time $t$. In single-target short-term tracking scenario there is only one object ($N_{t}^G = 1$, $SW_t = 0$) whose location can and should always be determined ($FP_t = 0$, $MI_t \in \{0, 1\}$), which means that the MOTA measure can be simplified to the percentage of correctly tracked frames, defined in equation (\ref{eq:ppositive}) earlier in this section.

\subsection{Tracking length}
\label{failure}

\begin{figure}[h]
    \centering
    \includegraphics[width = 0.9\columnwidth]{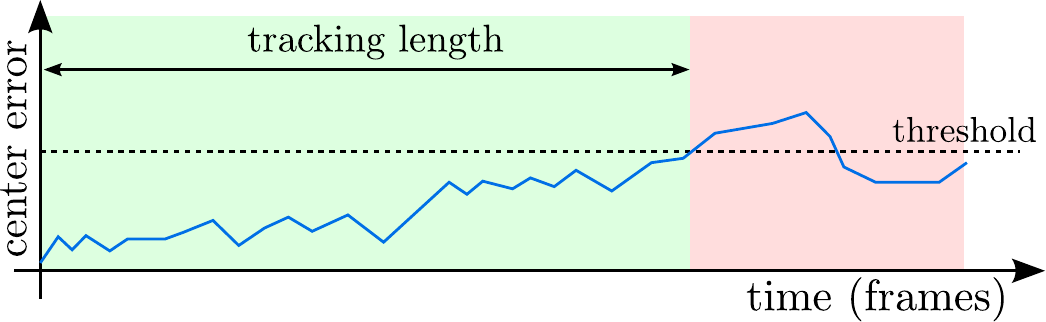}
    \caption{An illustration of the tracking length measure for center error. }
    \label{fig:failure}
\end{figure}

Another measure that has been used in the literature to compare trackers is {\em tracking length}~\cite{Kwon2009,Yang2014a}. This measure reports the number of successfully tracked frames from tracker's initialization to its (first) failure. A failure criterion can be a manual visual inspection (e.g. \cite{Godec2011}), which is biased and cannot be repeated reliably even by the same person. A better approach is to automate the failure criterion, e.g., by placing a threshold $\tau$ on the center or the overlap measure (see Figure~\ref{fig:failure}). The choice of the criterion may impact the result of comparison. As the overlap based criterion is more robust with respect to size changes, we will from now on denote in the following the tracking length measure with an overlap-based failure criterion by $L_{\tau}$ .

While this measure explicitly addresses the tracker's failure cases, which the simple average center-error and overlap measures do not, it suffers from a significant drawback. Namely, it only uses the part of the video sequence up to the first tracking failure. If by some coincidence, the beginning of the video sequence contains a difficult tracking situation, or the target is not visible well, which results in a necessarily poor initialization, the tracker will fail, and the remainder of the sequence will be discarded. This means that, technically, one would require a significant amount of sequences exhibiting the various properties right at its beginning to get a good statistic on this performance measure.

\subsection{Failure rate}
\label{supervised}

\begin{figure}[h]
    \centering
    \includegraphics[width = 0.9\columnwidth]{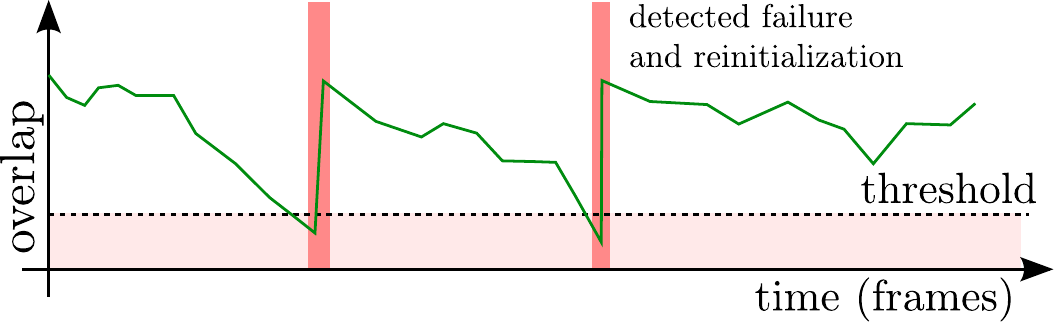}
    \caption{An illustration of the failure rate measure for overlap distance. }
    \label{fig:supervised}
\end{figure}

A measure that largely addresses the problem of the tracking length measure is the so-called {\em failure rate measure}~\cite{Kristan2010a,Khan2005}. The failure rate measure casts the tracking problem as a supervised system in which a human operator reinitializes the tracker once it fails. The number of required manual interventions per frame is recorded and used as a comparative score. The approach is illustrated in Figure~\ref{fig:supervised}. This measure also reflects the trackers performance in a real-world situation in which the human operator supervises the tracker and corrects its errors. Note that this performance measure should not be confused with different initialization strategies that can not be used as performance measures themselves (e.g. in~\cite{Wu2013} a tracker is initialized at different uniformly distributed positions in a sequence or with a random perturbation of the initialization region).

Compared to the tracking length measure, the failure rate approach has the advantage that the entire sequence is used in the evaluation process and decreases the importance of the beginning part of the sequence. The question of a failure criterion threshold is even more apparent here as each change in the criterion requires the entire experiment to be repeated. Researchers in \cite{Cehovin2011,Cehovin2013} consider a failure when the bounding box overlap is lower than $0.1$. This lower threshold is reasonable for non-rigid objects, since these are often poorly described by the bounding-box area. An even lower threshold could be used for overlap-based failure criteria if we are interested only in the most apparent failures with no overlap between the regions. We will denote the failure rate measure with an overlap-based failure criterion with threshold $\tau$ as

\begin{equation}
\label{eq:failurerate}
F_{\tau} = | \mathcal{F}_{\tau} |, \ \ \ \mathcal{F}_{\tau} = \{ f_i \},
\end{equation}

\noindent where $\mathcal{F}_{\tau}$ denotes the set of all failure frame numbers $f_i$. A drawback of the failure rate is that it does not reflect the distribution of these failures across the sequence. A tracker may fail uniformly in approximately equal intervals or it may fail more frequently at certain events. We can analyze these different distributions by looking at the {\em fragmentation} of the trajectory that is caused by the failures. Using an information theoretic point of view~\cite{Shannon2001}, we define the following trajectory fragmentation indicator, $\textrm{Fr}(\mathcal{F}_{\tau})$,

\begin{multline}
\label{eq:fragmentation}
\textrm{Fr}(\mathcal{F}_{\tau}) = \frac{1}{\log F_{\tau}} \sum_{f_i \in \mathcal{F}_{\tau}} -\frac{\Delta f_i}{N} \log \frac{\Delta f_i}{N}, \\  \Delta f_i = \begin{dcases*}
        f_{i+1} - f_i & when $f_i < \max(\mathcal{F}_{\tau})$\\
        f_1 + N - f_i & when $f_i = \max(\mathcal{F}_{\tau})$ 
        \end{dcases*},
\end{multline}

\noindent where $F$ denotes the number of failures and $f_i$ denotes the position of the $i$-th failure. The special case for the last failure ensures that the resulting value is not distorted by the beginning and end of the sequence\footnote{We interpret the sequence as a circular time-series and join the first and the last fragment. This way the value of $Fr$ stays the same for the shifts of the same distribution of failures.}. Fragmentation is only meaningful when $| \mathcal{F}_{\tau} | > 1$ as we are observing the inter-failure intervals. Maximum value $1$ is reached when the failures are uniformly distributed over the sequence and the value decreases when the inter-failure intervals become unevenly distributed. Note that the fragmentation can only be used as a supplementary indicator to the failure rate since it contains only limited information about the performance of a tracker, e.g. it will produce the same value for trackers that fail uniformly throughout the sequence no matter how many times they fail. However, it can be used to discriminate between trackers that fail frequently at a specific interval and those that fail uniformly over the entire sequence. As the evaluation datasets are getting larger, additional scores like fragmentation can help interpreting results on a higher level which we will demonstrate in Section~\ref{result}.

\subsection{\label{hybrid}Hybrid measures}

\begin{figure}[h]
    \centering
    \includegraphics[width = \columnwidth]{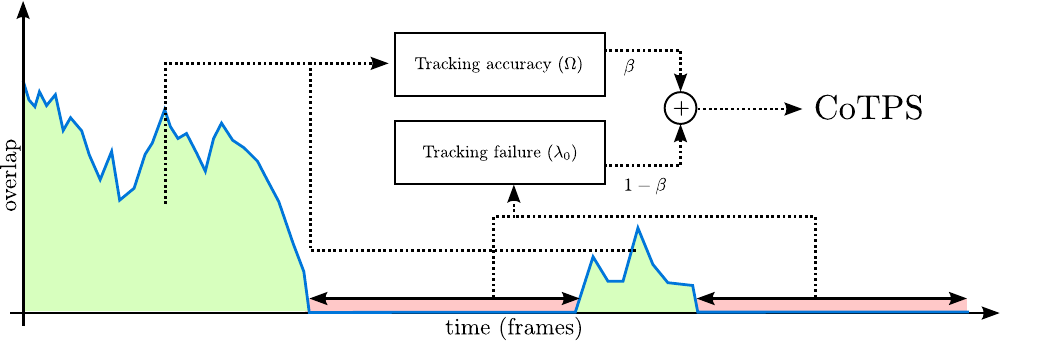}
    \caption{An illustration of the the CoTPS measure as described in~\cite{Nawaz2012}.}
    \label{fig:cotps}
\end{figure}

Nawaz and Cavallaro~\cite{Nawaz2012} propose a threshold-independent overlap-based measure that combines the information on tracking accuracy and tracking failure into a single score. This hybrid measure is called the {\em Combined Tracking Performance Score} (CoTPS) and is defined as a weighted sum of an accuracy score and a failure score. High score indicates poor tracking performance. The intuition behind CoTPS is illustrated in Figure \ref{fig:cotps}. At a glance, an appealing property of this measure is that it orders trackers by accounting for two separate aspects of tracking. However, no justification, neither theoretical nor experimental, is given of such rather complicated fusion which makes interpretation of this measure rather difficult. It can be shown (see Appendix~\ref{ap:cotps}) that the CoTPS measure can be reformulated in terms of average overlap, $\bar{\phi}$, and percentage of failure frames (where overlap is $0$), $\lambda_0$, i.e.

\begin{equation}
\label{eq:cotps-re}
CoTPS = 1 - \bar{\phi} - (1 - \lambda_0) \lambda_0.
\end{equation}

The equation (\ref{eq:cotps-re}) conclusively states that two very different basic measures are being combined in a rather complicated manner, prohibiting a straightforward interpretation.
Precisely, if one tracker is ranked higher than another one it is not clear if this is due to a higher average overlap or less failed frames. Furthermore, if equation (\ref{eq:cotps-re}) is reformulated as $CoTPS = (1 - \lambda_0) ( 1 - \hat{\phi}) + \lambda_0^2$, where the $\hat{\phi}$ denotes the average overlap on non-failure frames (where the overlap is greater than $0$), multiple combinations of two values produce the same CoTPS score. In Figure~\ref{fig:cotps-equal} we illustrate several such equality classes, where the same CoTPS score is achieved using different combinations of the two components, which makes the interpretation of the results difficult.
The combined score is also inconvenient in scenarios where a different combination of performance properties is desired.

In terms of performance score, we therefore believe that a better strategy is to focus on a few complementary performance measures with a well-defined meaning, and avoid fusing them into a single measure early on in the evaluation process.

\subsection{\label{plots}Performance plots}

\begin{figure*}[t!]
\begin{minipage}[b]{0.33\linewidth}
    \centering
    \includegraphics[width = 0.8\textwidth]{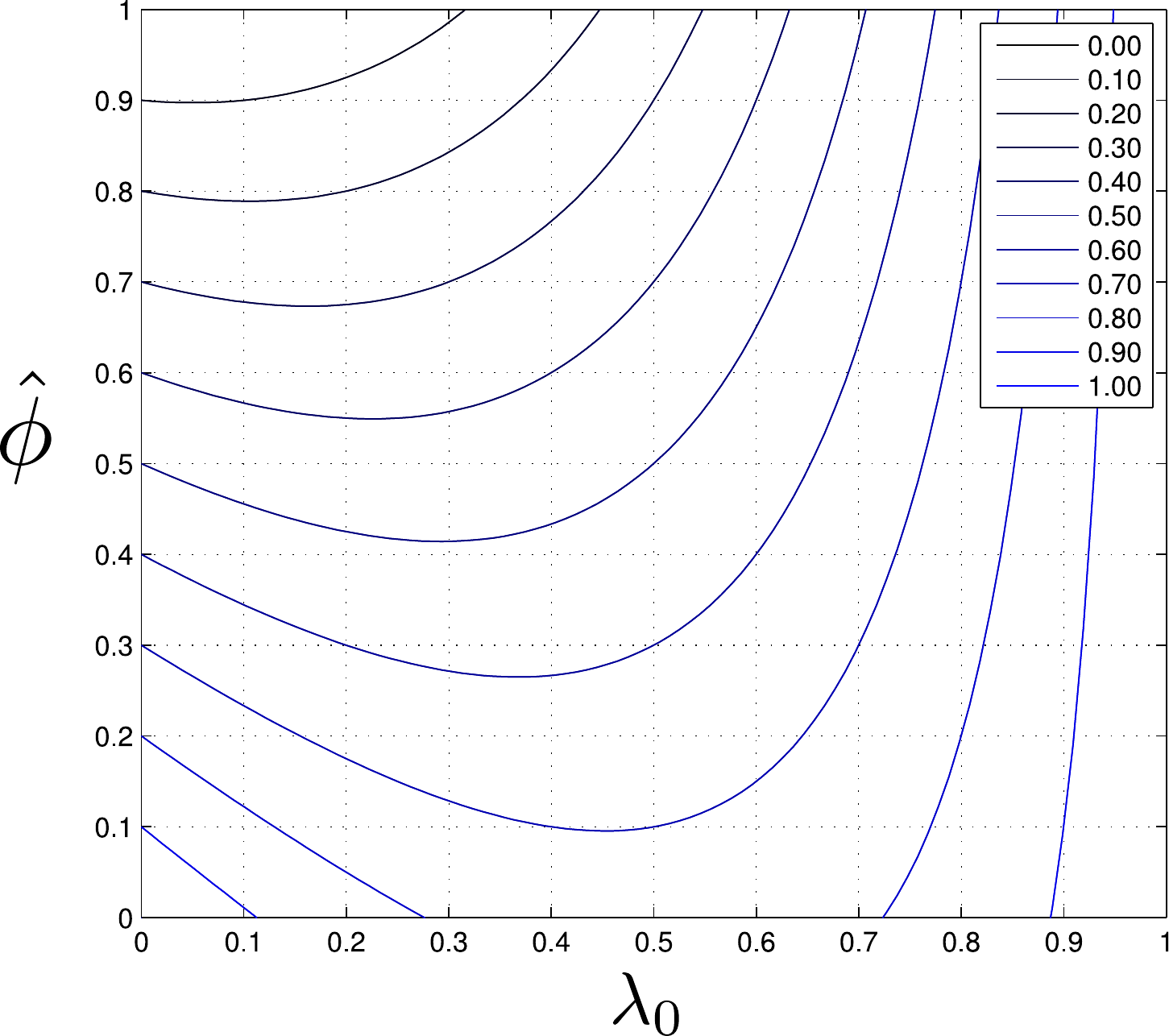}
    \caption{Equality classes for different values of CoTPS measure. Each line denotes pairs of average overlap on non-failed frames ($\hat{\phi}$) and percentage of failure frames ($\lambda_0$) that produce the same CoTPS score.}
    \label{fig:cotps-equal} 
\end{minipage}
\hspace{0.1cm}
\begin{minipage}[b]{0.36\linewidth}
    \centering
    \includegraphics[width = \textwidth]{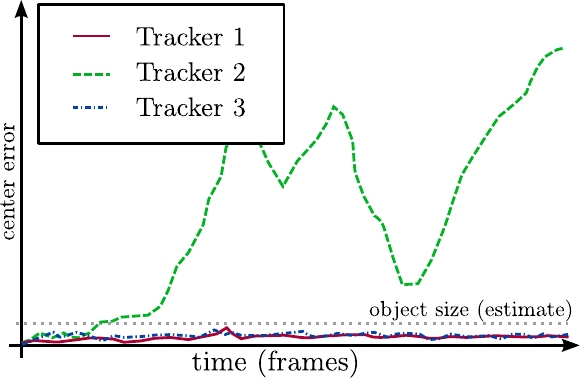}
    \caption{An example of center-error plot comparison for three trackers. Tracker $2$ has clearly failed in the process, yet its large center errors cause the plot to expand its vertical scale, thus reducing the apparent differences of trackers $1$ and $3$.}
    \label{fig:errorplot}
\end{minipage}
\hspace{0.1cm}
\begin{minipage}[b]{0.33\linewidth}
    \centering
    \includegraphics[width = 0.8\textwidth]{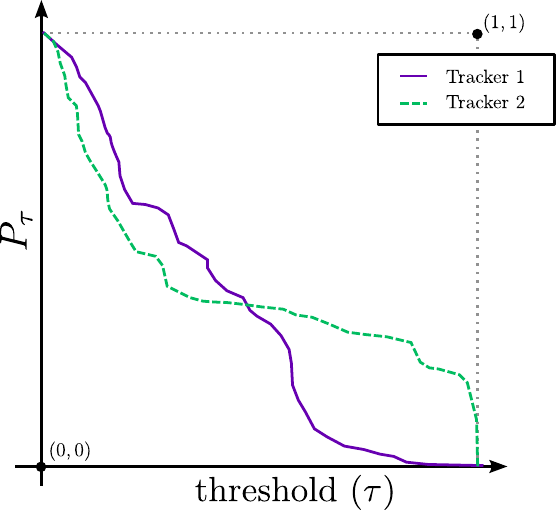}
    \caption{An illustration of the {\em measure-threshold} plot for two trackers. It is apparent that different values of the threshold would clearly yield different order of the trackers.}
    \label{fig:overlaproc} 
\end{minipage}
\end{figure*}

Plots are frequently used to visualize the behavior of a tracker since they offer a clearer overview of performance when considering multiple trackers or sets of tracker parameters. The most widely-used plot is a center-error plot that shows the center-error with respect to the frame number~\cite{Babenko2011,Adam2006,Bao2012,Zhang2012}. While this kind of plots can be useful for visualizing tracking result of a single tracker, a combined plot for multiple trackers is in many cases misused if applied without caution, because the tracker with an inferior performance ``steals away'' the focus from the information that we are interested in with this type of plots, i.e. the tracker accuracy. An illustration of such a problematic plot is shown in Figure \ref{fig:errorplot} where two trackers appear equal due to a distorted scale caused by the third tracker. A less popular but better bounded alternative approach is to plot region overlap, e.g. in~\cite{Wang2013}.

In the previous section we have seen that a failure criterion plays a significant role in visual tracker performance evaluation. Choosing an appropriate value for the threshold may affect the order and can also be potentially misused to influence the results of a comparison. However, it is sometimes better to avoid the use of a single specific threshold altogether, especially when the evaluation goal is general and a specific threshold is not a part of the target task. To avoid the choice of a specific threshold, results can be presented as a {\em measure-threshold} plot. This kind of plots have some resemblances to a ROC curve~\cite{Fawcett2006}, like monotony, intuitive visual comparison, and a similar calculation algorithm. Measure-threshold plots were used in~\cite{Babenko2011}, where the authors used center-error as a measure as well as in~\cite{Wu2013}, where both center-error and overlap are used.

The percentage of correctly tracked frames, defined in (\ref{eq:ppositive}) as $P_{\tau}$, is a good choice for a measure to be used in this scenario, however, other measures could be used as well. The $P_{\tau}$ measure can be intuitively computed for multiple sequences which makes it useful for summarizing the entire experiment (an example of $P_{\tau}$ plot is illustrated in Figure~\ref{fig:overlaproc}). Interpretations of such plots have been so far limited to their basic properties which in a way negates the information verbosity of a graphical representation. For example, similarly to ROC curves, we can compute an area-under-the-curve (AUC) summarization score, which is used in~\cite{Wu2013,Nawaz2012} to reason about the performance of the trackers. However, the authors of~\cite{Wu2013,Nawaz2012} do not provide an interpretation of this score. We prove  in this paper (see Appendix~\ref{ap:auc}) that the AUC is in fact the average overlap, which results in two important implications: (1) the complicated computation of ROC-like curve and subsequent numerical integration for calculating AUC can be avoided by simple averaging of overlap over the sequence and (2) the AUC has a straight-forward interpretation. 

A curve that is visually similar to $P_{\tau}$ plot is the {\em survival curve}~\cite{Smeulders2013}. In this case the curve summarizes the trackers' success (various performance measures can be used) over a dataset of sequences that are ordered from the best performance to the worst. While this approach gives a good overview of the overall success, it is not suitable for a sequence-wise comparison as the order of sequences differs from tracker to tracker. Not all sequences are equal in terms of difficulty as well as in terms of the phenomena that they contain (e.g. occlusion, illumination changes, blur) which makes it very hard to interpret the results of a survival curve on a more detailed level.

\section{Experimental comparison of performance measures}
\label{result}

The theoretical analysis so far shows that different measures may reflect different aspects of tracking performance, so it is impossible to simply say which the best measure is. Furthermore some measures are proven to be equal (e.g., area-under-the-curve and average overlap). We start our analysis by establishing similarities and equivalence between various measures, by experimentally analyzing which measures produce consistently similar responses in tracker comparison. The main idea is that strongly correlated measures are sensitive the same quality of a visual tracker, therefore we should only consider a subset of measures that are not correlated or at most weakly correlated.

In order to analyze the performance measures, we have conducted a comparative experiment. Our goal is to evaluate several existing trackers according to the selected measures on a number of typical visual tracking sequences. The selection of measures is based on our theoretical discussion in Section~\ref{measures}. We have selected the following measures: 
\begin{enumerate}
\item average center error (Section~\ref{distance}),
\item average normalized center error (Section~\ref{distance}),
\item root-mean-square error (Section~\ref{distance}),
\item percent of correct frames for $\tau = 0.1$, $P_{0.1}$ (Section~\ref{overlap}),
\item percent of correct frames for $\tau = 0.5$, $P_{0.5}$,
\item tracking length for threshold $\tau > 0.1$, $L_{0.1}$ (Section~\ref{failure}),
\item tracking length for threshold $\tau > 0.5$, $L_{0.5}$,
\item average overlap (Section~\ref{overlap}),
\item Hybrid CoTPS measure (Section~\ref{hybrid}),
\item average center error for $F_0$,
\item average normalized center error for $F_0$,
\item root-mean-square error for $F_0$,
\item percent of correct frames for $\tau = 0.1$, $P_{0.1}$ for $F_0$,
\item percent of correct frames for $\tau = 0.5$, $P_{0.5}$ for $F_0$,
\item average overlap in case of $F_{0}$,
\item failure rate $F_{0}$ (Section~\ref{supervised}).
\end{enumerate}

The first nine measures were calculated on trajectories where the tracker was initialized only at the beginning of the sequence, and the remaining seven measures were calculated on trajectories where the tracker was reinitialized if the overlap between predicted and ground-truth region became $0$.

Since the goal of the experiment is not evaluation of trackers but selection of measures, the main guideline when selecting trackers for the experiment was to create a diverse set of tracking approaches that fail in different scenarios and are therefore capable of showing differences of evaluated measures on real tracking examples. We have selected a diverse set of $16$ trackers, containing various detection-based trackers, holistic generative trackers, and part-based trackers, that were proposed in the recent years: A color-based particle filter (PF)~\cite{Perez2002}, the On-line boosting tracker (OBT)~\cite{Grabner2006}, the Flock-of-features tracker (FOF)~\cite{Kolsch2004}, the Basin-hopping Monte Carlo tracker (BHMC)~\cite{Kwon2009}, the Incremental visual tracker (IVT)~\cite{Ross2008}, the Histograms-of-blocks tracker (BH)~\cite{Nejhum2010}, the Multiple instance tracker (MIL)~\cite{Babenko2011}, the Fragment tracker (FRT)~\cite{Adam2006}, the P-N tracker (TLD)~\cite{Kalal2010}, the Local-global tracker (LGT)~\cite{Cehovin2013}, Hough tracker (HT)~\cite{Godec2011}, the L1 Tracker Using Accelerated Proximal Gradient Approach (L1-APG)~\cite{Bao2012}, the Compressive tracker (CT)~\cite{Zhang2012}, the Structured SVM tracker (STR)~\cite{Hare2011}, the Kernelized Correlation Filter tracker (KCF)~\cite{Henriques2014}, and the Spatio-temporal Context tracker (STC)~\cite{Zhang2014}. The source code of the trackers was provided by the authors and adapted to fit into our evaluation framework. 

\begin{figure}[h]
    \centering
    \includegraphics[width = \columnwidth]{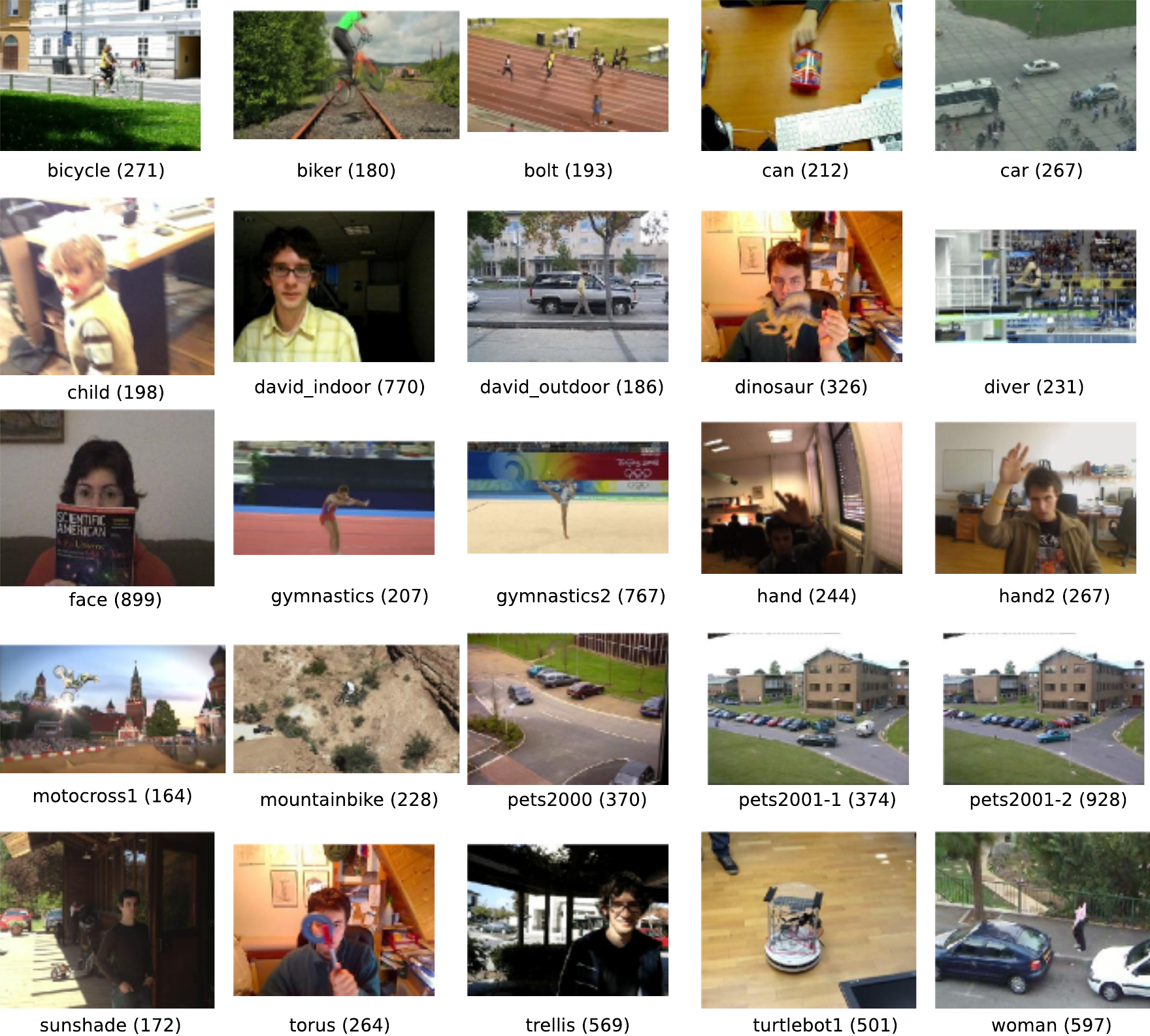}
    \caption{Overview of the sequences used in the experiment. The number in brackets besides the name denotes the length of a sequence in frames.}
    \label{fig:sequences}
\end{figure}

We have run the trackers on $25$ different sequences, most of which are well-known in the visual tracking community, e.g.~\cite{Cehovin2013,Cehovin2011,Zhang2012,Wang2011,Ross2008,Kwon2009,Adam2006,Godec2011}, and several were acquired additionally. Representative images from the sequences are shown in Figure~\ref{fig:sequences}. The sequences were annotated with an axis-aligned bounding-box region of the object (if the annotations were not already available), as well as the central point of the object, in cases where the center of the object did not match the center of the bounding-box region.
To account for stochastic processes that are a part of many trackers, each tracker was executed on each sequence $30$ times. The parameters for all trackers were set to their default values and kept constant during the experiment. A separate run was executed for the {\em failure rate} measure as the re-initialization influences other aspects of tracking performance. 

Because of the scale of the experiment, only the most relevant results are presented in Section \ref{result}. Additional results, such as the ordering of the trackers according to individual measures, are available in the supplementary material\footnote{Supplementary material is available at \url{http://go.vicos.si/performancemeasures}.}.

\subsection{Correlation analysis}
\label{sec:correlation}
\begin{figure}[h]
    \centering
    \includegraphics[width = \columnwidth]{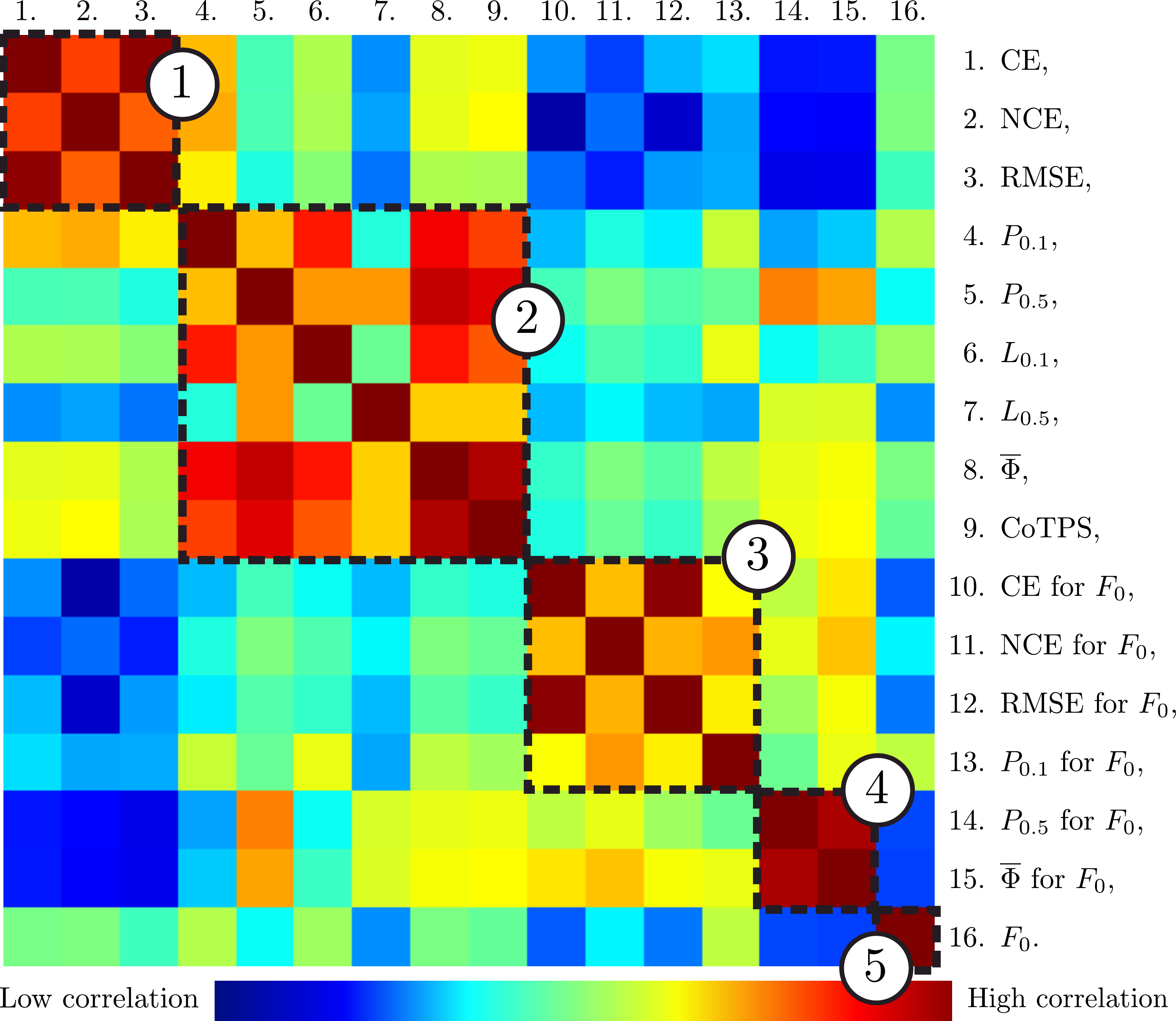}
    \caption{Correlation matrix for all measures visualized as a heat-map overlaid with obtained clusters. The image is best viewed in color.}
    \label{fig:correlation}
\end{figure}

A correlation matrix was computed from all pairs of measures calculated over all tracker-sequence pairs. Note that we do not calculate the correlation on rankings to avoid handling situations where several trackers take the same place (if differences are not statistically significant). The rationale is that strongly correlated measure values will also produce similar order for trackers. Since we have run $16$ trackers, each of the stochastic ones was run $30$ times on every sequence, this means that every performance measure has about $10000$ samples. This is more than enough for statistical evaluation of whether correlation across the measures exists. The obtained correlation matrix is shown in Figure \ref{fig:correlation}. Using automatic cluster discovery by affinity propagation~\cite{Frey2007} we have determined five distinct clusters, one for measures $1$ to $3$, one for measures $4$ to $9$, one for measures $10$ to $13$, one for measures $14$ and $15$, and one for measure $16$. All these correlations are highly statistically significant ($p < 0.001$).

\begin{figure}[h]
    \centering
    \includegraphics[width = \linewidth]{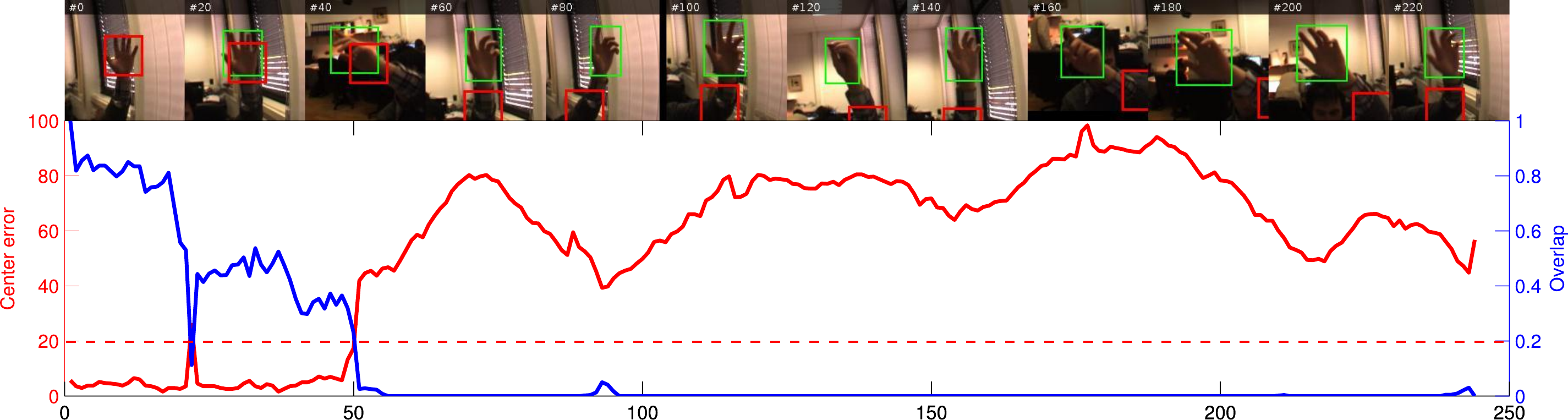}
    \caption{A comparison of {\em overlap} and {\em center error distance} measures for tracker CT on sequence {\em hand}~\cite{Cehovin2011}. The dashed line shows the estimated threshold above which the center error is greater than the size of the object. The tracker fails around frame $50$. }
    \label{fig:overlap_vs_error}
\end{figure}

The first cluster of measures consists of the three center-error-based measures. This is expected since all of these measures are based on {\em center-error} using different averaging methods. The second cluster of measures contains {\em average overlap}, {\em percentage of correctly tracked frames} for two threshold values ($P_{0.1}$ and $P_{0.5}$) and {\em tracking length} ($L_{0.1}$ and $L_{0.5}$). Measures in the second cluster assume that incorrectly tracked frames do not influence the final score based on the specific (incorrect) position of the tracker. Because of this and the insensitivity to the scale changes they are a better choice to measure  tracking performance than the center-error-based measures. An illustration of this difference for {\em overlap} and {\em center-error} is shown as a graph in Figure \ref{fig:overlap_vs_error}, where we can clearly see that the center-error measure takes into account the exact center distance at frames after the failure has occurred, which depends on the movement of an already failed tracker and does not reflect its true performance.

\begin{figure}[h]
    \centering
    \includegraphics[width = \linewidth]{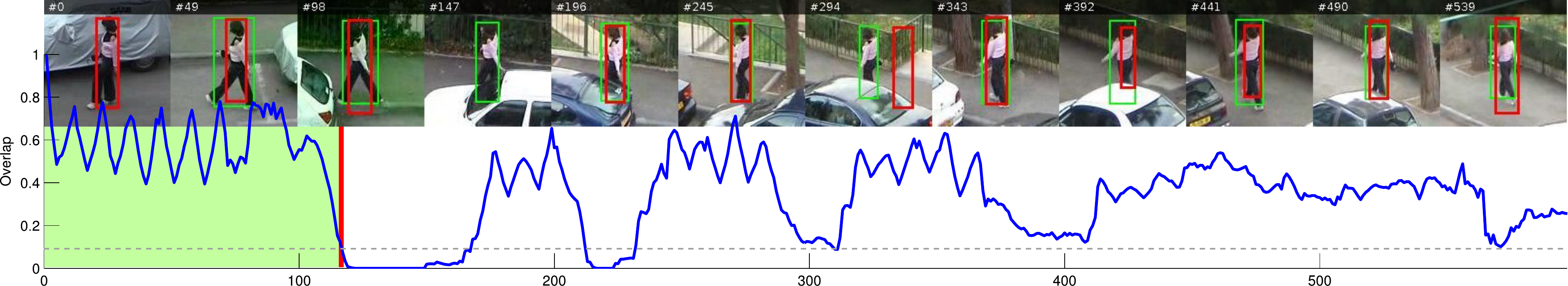}
    \caption{An overlap plot for tracker TLD on sequence {\em woman}~\cite{Adam2006}. The dashed line shows the threshold below which the tracking length detects failure (for threshold $0.1$), which happens around frame $120$. }
    \label{fig:overlap_vs_length}
\end{figure}

The first cluster of measures in Figure~\ref{fig:correlation} implies that the first three measures are equivalent and it does not matter which one is chosen. The second cluster requires further interpretation. Despite the apparent similarity of overlap-based measures $4$ to $8$ and of the CoTPS measure, the correlation is not perfect and the order of trackers differ in some cases. One example of such a difference can be seen for the TLD tracker on the {\em woman} sequence (Figure \ref{fig:overlap_vs_length}). We can see that the tracker loses the target early on in the sequence (during an occlusion), but manages to locate it again later because of its discriminative nature. The {\em average overlap} (Measure $8$) and the {\em percentage of correct frames} (Measures $4$ and $5$) therefore order the tracker higher than the {\em tracking length} (Measures $6$ and $7$). On the general level we can also observe that the choice of a threshold can influence the outcome of the experiment. This can be observed for tracking length measures $L_{0.1}$ and $L_{0.5}$ and to some extent for the percentage of correct frames measures $P_{0.1}$ and $P_{0.5}$. In those cases, the scores for a higher threshold ($0.5$) result in a different order of trackers compared to the lower threshold ($0.1$). This means that care must be taken when choosing the thresholds as they may affect the outcome of the evaluation. While a certain threshold may be given for a specific application domain, it is best to avoid it in general performance evaluation. The last measure in the second cluster is the hybrid CoTPS measure~\cite{Nawaz2012} which turns out to be especially strongly correlated with the average overlap measure. By looking back at our theoretical analysis in Section~\ref{hybrid} the CoTPS produces identical results for trajectories where the overlap never reaches $0$ (no failure). In other cases the percent of failed frames, which can be approximated using $1 - P_{0.1}$, is also strongly correlated with average overlap. This means that the entire measure is biased towards only one aspect of tracking performance.

We can in fact observe a slight overlap between the first two clusters in the correlation matrix, implying similarity in their information content. Based on the above analysis and discussion in Section \ref{measures} we conclude that the {\em average overlap} measure is the most appropriate to be used in tracker comparison, as it is simple to compute, it is scale and threshold invariant, exploits the entire sequence, and it is easy to interpret. Note also that it is highly correlated with a more complex {\em percentage-of-correctly-tracked-frames} measure.

\begin{figure}[h]
    \centering
    \includegraphics[width = \linewidth]{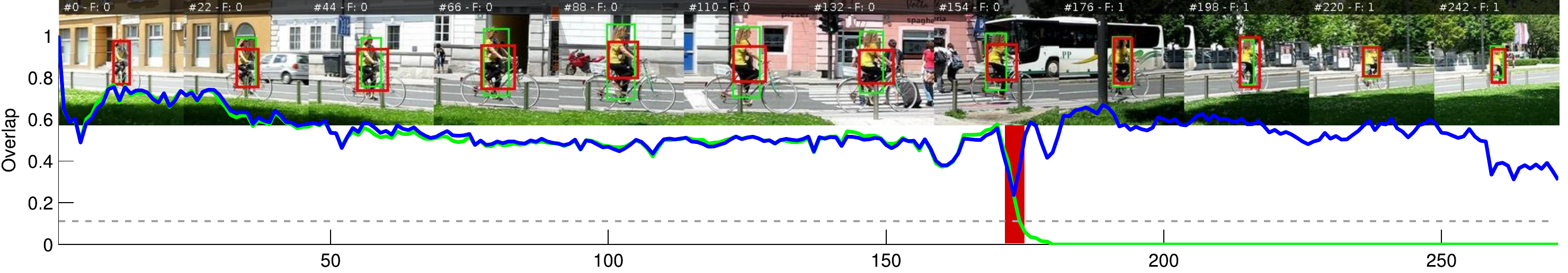}
    \caption{An overlap plot for tracker LGT on sequence {\em bicycle}~\cite{Cehovin2013}. The green plot shows the unsupervised overlap, and the blue plot shows the overlap for supervised tracking, where the failure is recorded and the tracker re-initialized.}
    \label{fig:overlap_vs_failure}
\end{figure}

The {\em failure rate} measure influences the trackers' entire trajectory, because of the re-initializations. The data for measures $9$ to $16$ was therefore acquired as a separate experiment. The advantage of the supervised tracking scenario is that the entire sequence is used, which makes the results statistically significant at smaller number of sequences. It does not matter that much if one tracker fails at the ``difficult'' beginning of the sequence, while the other one barely survives and then tracks the rest successfully. While supervised evaluation looks more complex, this is a technical issue that can be solved with standardization of evaluation process~\cite{Cehovin2014a}. In Figure \ref{fig:overlap_vs_failure} we can see the performance of the LGT tracker on the {\em bicycle} sequence. Because of a short partial occlusion near frame $175$ the tracker fails, although it is clearly capable of tracking the rest of the sequence reliably if re-initialized. Measures that are computed on the trajectories with reinitialization exhibit similar correlation relations than for the trajectories without reinitialization.

According to the correlation analysis the least correlated measures are failure rate and average overlap on re-initialized trajectories. These findings are discussed in next section where we propose a conceptual framework for their joint interpretation. To further support the stability of the measurements, we have also performed the correlation analysis on different subsets of approximately half of the total $25$ sequences and found that the these findings do not change.

\subsection{Accuracy vs. robustness} 
% TODO: omeni http://mi.eng.cam.ac.uk/~bdrs2/papers/stenger_cvpr09.pdf
An intuitive way to present tracker performance is in terms of accuracy (i.e., how accurately the tracker determines the position of the object) and robustness (i.e., how many times the tracker fails). Based on the correlation analysis in Section~\ref{sec:correlation} we have selected a pair of evaluated measures that estimates the aforementioned qualities. The {\em average overlap} measure is the best choice for measuring the accuracy of a tracker because it takes into account the size of the object and does not require a threshold parameter. However, it does not tell us much about the robustness of the tracker, especially if the tracker fails early in the sequence. The {\em failure rate} measure, on the other hand, measures the number of the failures which can be interpreted as robustness of the tracker. According to correlation analysis in Section~\ref{sec:correlation}, if we measure average overlap on the re-initialized data, used to estimate failure rate, the measures are not correlated. This is a desired property as they should measure different aspects of tracker performance. We thus propose measuring the short-term tracking performance by the following A-R pair,

\begin{equation}
\label{eq:armeasure}
\textrm{A-R}(\Lambda^G, \Lambda^T) = \left( \overline{\Phi}(\Lambda^G, \Lambda^T), F_0(\Lambda^G, \Lambda^T) \right),
\end{equation}
\noindent where $\overline{\Phi}$ denotes {\em average overlap} and $F_0$ denotes the {\em failure rate} for $\tau = 0$. Note that the value of failure threshold $\tau$ can influence the final results. If the value is set to a high value (i.e. close to $1$) the tracker is restarted frequently even for small errors and the final score is hard to interpret. Based on our analysis, we propose to use the lowest theoretical threshold $\tau = 0$ to only measure complete failures where the regions have no overlap at all and a reinitialization is clearly justified. In theory, a tracker can also report an extremely large region as the position of the target and avoids failures, however, the accuracy will be very low in this case. This is an illustrative example of how the two measures complement each other in accurately describing the tracking performance.

It is worth noting that there are some parallels between the hybrid CoTPS measure~\cite{Nawaz2012}, and the proposed A-R measure pair. In both cases two aspects of tracking performance are considered. The first part of the CoTPS measure is based on the AUC of the overlap plot, which, as we have shown, is equal to average overlap. The second part of the measure attempts to report tracker failure by measuring the number of frames where the tracker has failed (overlap is $0$), which could also be written as $P_{0}$. Despite these apparent similarities, the A-R measure pair is better suited for visual tracker evaluation for several reasons: (1) the chosen measures are not correlated, (2) the supervised evaluation protocol uses sequences more effectively because of reinitializations, (3) different performance profiles for average overlap and failure rate produce different combinations of scores that can be interpreted, which is not true for CoTPS measure. 

\begin{figure}[h]
    \centering
    \includegraphics[width = 0.9\linewidth]{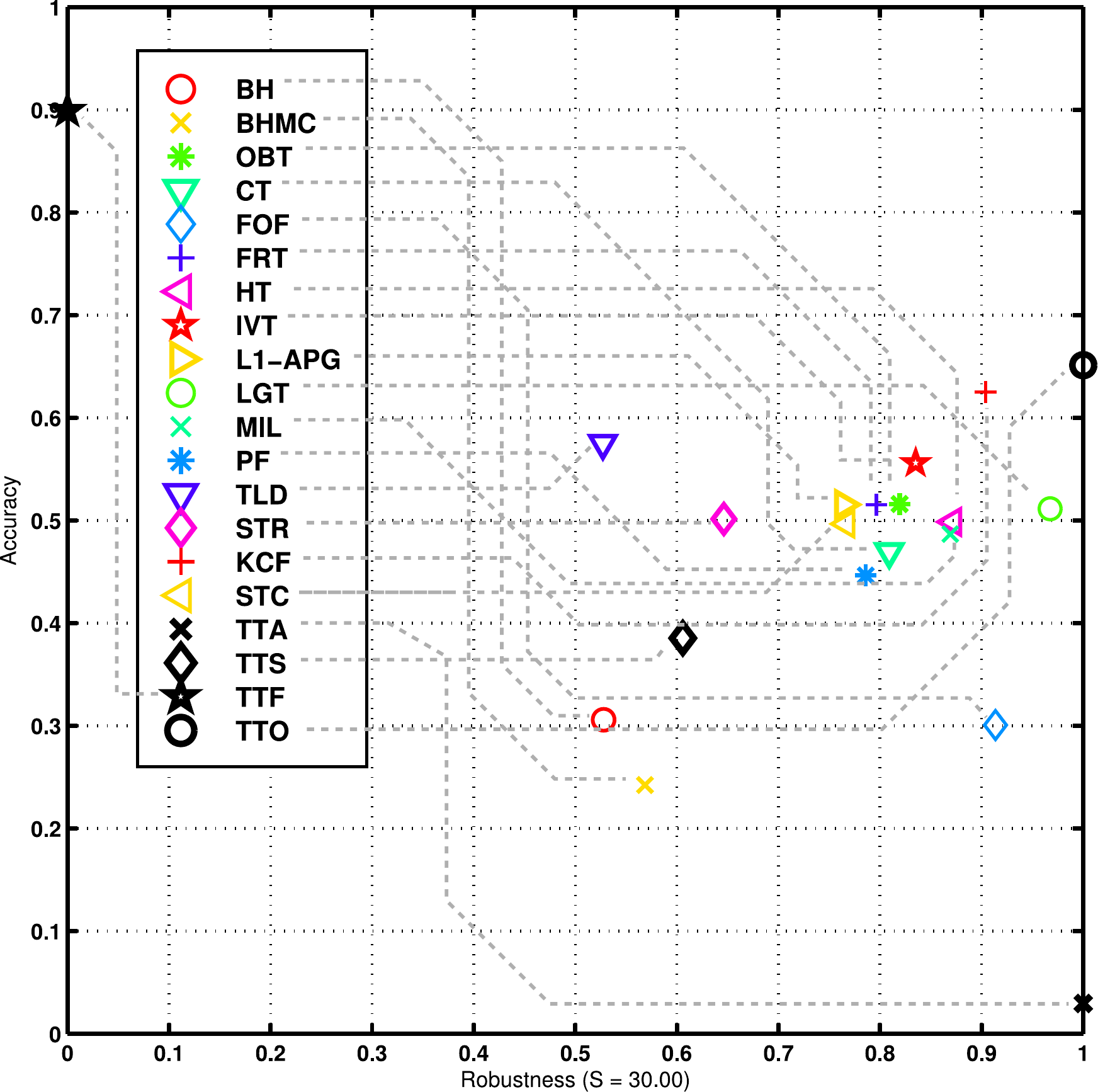}
    \caption{An accuracy-reliability data visualization for all trackers over all sequences.}
    \label{fig:ar_plot}
\end{figure} 

A pair of measures is most efficiently represented via visualization. We propose to visualize the A-R pair as a $2$-D scatter plot. This kind of visualization is indeed very simple, but is easy to interpret, extendable and has been used in visual tracking visualization before, e.g.~\cite{Stenger2009}. An example of an A-R plot for the data from the experiment can be seen in Figure~\ref{fig:ar_plot}, where we show the average scores for all sequences, from which one can read the trackers performance in terms of accuracy (the tracker is more accurate if it is higher along the vertical axis) and robustness (the tracker fails fewer times if it is further to the right on the horizontal axis). Because the robustness does not have an upper bound we propose to interpret it as {\em reliability} for visualization purposes. The reliability of a tracker is defined as an exponential failure distribution, $R_S = e^{- S M }$. The value of $M$ denotes mean-time-between-failures, i.e. $M = \frac{F_0}{N}$, where $N$ is the length of the sequence. The reliability of a tracker can be interpreted as a probability that the tracker will still successfully track the object up to $S$ frames since the last failure, assuming a uniform failure distribution that does not depend on previous failures. This is of course not true in all cases, however, note that this formulation and the choice of $S$ does not influence the order of the trackers but can be adjusted as a scaling factor for better visualization. Interpreting results this way is useful for visualization and quick interpretation of results, however, one should still consult the detailed values of average overlap and failure rate before making any final decisions.

\subsection{Theoretical trackers}
\label{sec:theoretical}

For a better understanding of the complementing nature of the two measures we introduce four theoretical trackers denoting extreme prototypical tracker behaviors. The {\em first theoretical tracker}, denoted by TTA, always reports the region of the object to equal the image size of the sequence. This tracker provides regions that are too loose, but does not fail (overlap is never $0$) and is therefore displayed in the bottom-right corner as it is extremely robust, but not accurate at all. The {\em second theoretical tracker}, denoted by TTS, reports its initial position for the entire sequence. This tracker will likely fail if the object moves, and will achieve better accuracy because of frequent manual interventions. The {\em third theoretical tracker}, denoted by TTF only tracks one frame and then deliberately reports a failure. This way the tracker maintains a high accuracy, however the failure rate is extremely high and the tracker is placed in top-left corner of the plot. The {\em fourth theoretical tracker} is denoted as TTO and represents an {\em oracle} tracker of fixed size. The tracker always correctly predicts the center position of the object, however, the size of the object is fixed. This tracker represents a practical performance limit for trackers that do not adapt the size of the reported bounding box which is the same as the initialization bounding box.

The performance scores for the theoretical trackers can be easily computed directly from ground-truth. The simplicity, intuitive nature, and the parameter-less design make them excellent interpretation guides in the graphical representations of results, such as A-R plot. In other words, they put the results of evaluated trackers into context by providing reference points for a given evaluation sequence.

\subsection{Interpretation of results using A-R plots}

By establishing the selection of measures, visualization and the theoretical trackers as an interpretation guide, we can now provide an example of results interpretation. The A-R plot in Figure~\ref{fig:ar_plot} shows results, averaged over entire data-set. We can see that the LGT tracker is on average the most robust one in the set of evaluated trackers (positioned most right), but is surpassed in terms of accuracy by KCF, IVT and TLD (positioned higher). Espectially the TLD tracker is positioned very low in terms of robustness, so the high accuracy may in fact be a result of frequent reinitializations, a behavior that is similar to the TTF tracker. We acknowledge that this behavior of TLD is a design decision as the TLD is actually a long-term tracker that that does not report the position of the object if it is not certain about its location. The FOF tracker, on the other hand, is quite robust, but its accuracy is very low. This means that it most likely sacrifices accuracy by spreading accross a large portion of the frame, much like TTA.

As the averaged results can convey only a limited amount of information, we have also included per-sequence A-R plots in the supplementary material. These plots show that the actual performance of trackers differs significantly between the sequences. Theoretical trackers TTA and TTF remain worse on their individual axes as expected, while the relative position of the other trackers changes depending on the properties of the individual sequence. In many sequences the TTO tracker achieves the best performance because of its ability to ``predict'' the position of the target. In cases where the size of the object changes this advantage becomes less apparent and trackers like IVT, L1-APG, HT, and LGT that account for this change can even surpass it in terms of accuracy (e.g. in {\em biker}, {\em child}, and {\em pets2001-2}). The sequence {\em diver} is interesting considering the results. Even though the object does not move a lot in the image space, which is apparent from the high robustness of the TTS tracker, the sequence has nevertheless proven to be very challenging for most of the trackers because of the large deformations of the object. The BH and BHMC trackers are on average very similar to the TTS tracker which would mean that they do not cope well with moving objects. At a closer look we can see that this is only true for some sequences (e.g. {\em torus}, {\em bicycle}, and {\em pets2000}). In other sequences both tracker perform either better than TTS, where the background remains static and can be well separated from the object (e.g. {\em sunshade}, {\em david\_outdoor}, and {\em gymnastics2}), or worse, where the appearance of the background changes (e.g. {\em motocross1}, {\em child}, and {\em david\_indoor}). Considering the good average performance of the LGT tracker we can see that the tracker performed well in sequences with articulated and non-rigid objects (e.g. {\em hand}, {\em hand2}, {\em dinosaur}, {\em can}, and {\em torus}), while the difference in case of more rigid objects (e.g. {\em face}, {\em pets2001-1}, and {\em pets2001-2}) is less apparent. In the plot for the {\em bolt} sequence we can see that the TLD tracker behaves similarly to the TTF tracker, i.e. fails a lot without actually drifting. On the other hand the TLD tracker works quite well in the case of {\em pets2000}, {\em pets2001-1}, and {\em pets2001-2} sequences where the changes in the appearance of the object are gradual.

\subsection{Fragmentation} 

\begin{figure}[!t]
    \centering
    \includegraphics[width = \linewidth]{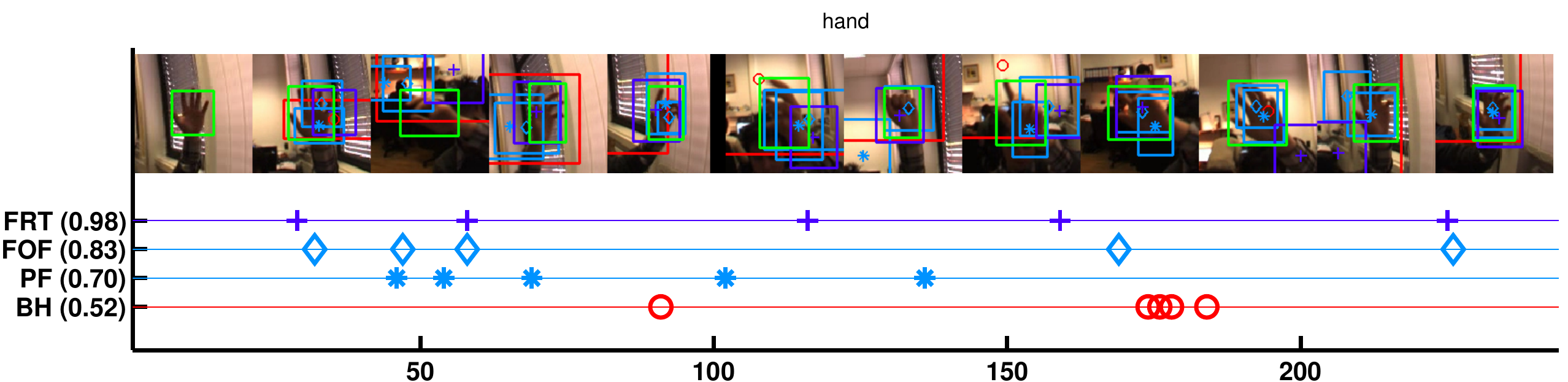}
    \includegraphics[width = \linewidth]{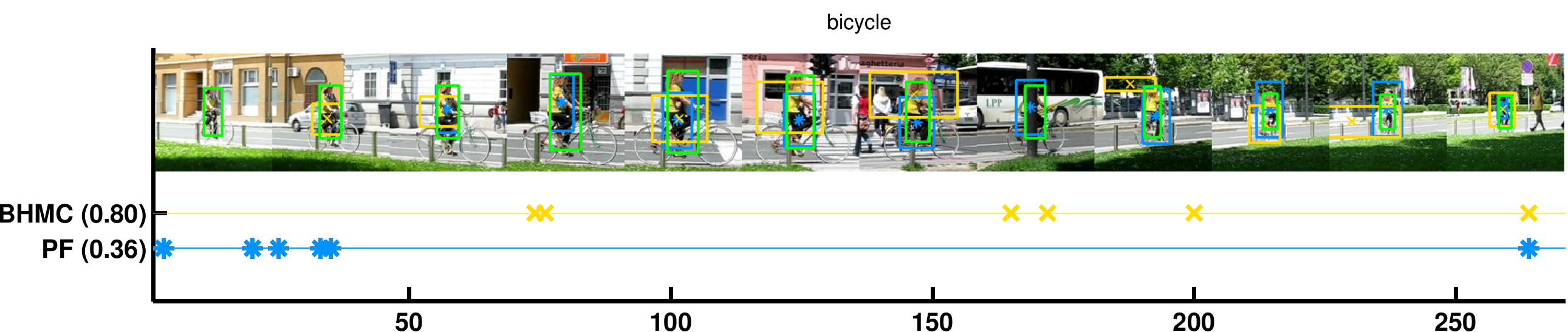}
    \includegraphics[width = \linewidth]{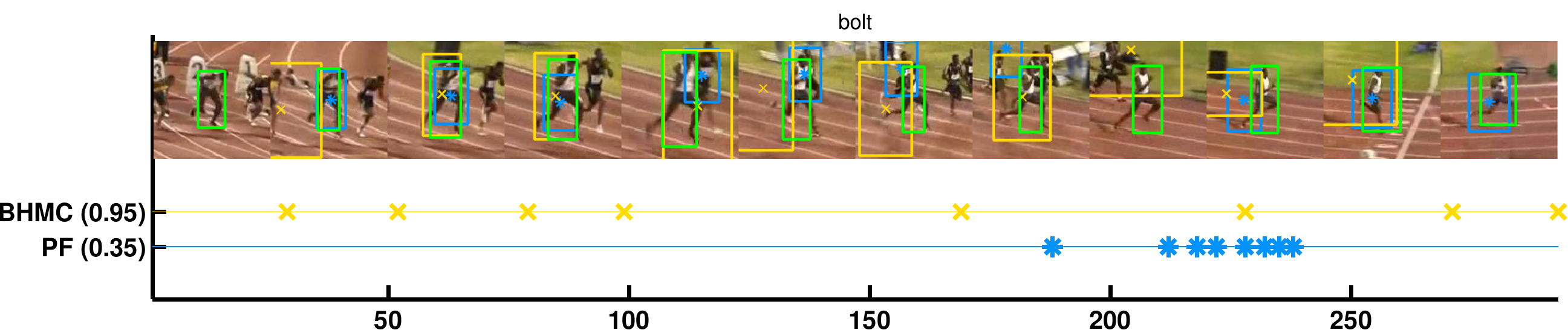}
    \caption{Selected results of the fragmentation analysis. Failures are marked on the time-line with symbols, the corresponding fragmentation values are shown in brackets next to tracker name. }
    \label{fig:fragmentation}
\end{figure}

Recall that we have introduced the fragmentation indicator as a complementary indicator for the number of failures measure in Equation~\ref{eq:fragmentation}. Using this measure we can infer some additional properties of a tracker that would otherwise require looking at raw results. Fragmentation reflects the distribution of failures throughout the sequence. If the fragmentation is low then the failures are likely clustered together around some specific event (which can indicate a specific event that is problematic for the tracker). On the other hand, if the fragmentation is high, then the failures are uniformly distributed, independently of localized events in the sequence and can be most likely attributed to internal problems of the tracker. To demonstrate this property we have selected several cases where the number of failures is the same, but the fragmentation is different. In Figure~\ref{fig:fragmentation} we can see three such cases. Several trackers, despite failing the same number of times do this for different reasons and in different intervals. On the {\em hand} sequence, the FRT tracker fails almost uniformly, while the BH tracker manages to hold to the target for a long time (the region is, however, estimated very poorly), but then fails to successfully initialize around frame $170$ because of background clutter and motion. In {\em bicycle} and {\em bolt} sequences, the failures of PF tracker are concentrated on a specific event, most likely because of color ambiguity or small target size. The failures of the BHMC tracker are almost uniformly distributed over both sequences, most likely because of the problems of the tracker implementation (e.g. inability to cope with small target size). 

\subsection{Sequences from perspective of theoretical trackers} 

The theoretical trackers, introduced in Section~\ref{sec:theoretical}, provide further insights into each sequence from the perspective of the basic properties that each theoretical tracker represents. Because of their simplicity and absence of parameters, they can easily be applied to any annotated sequence and provide some insight about its properties. These properties can then be used when constructing an evaluation dataset or interpreting the results. 

The TTA tracker will always achieve good robustness (no failures), but will produce high accuracy values only when the target will cover large part of the image frame. This tracker therefore measures the average relative size of the object. The TTS tracker will only achieve good robustness when the object remains stationary with respect to the image plane (e.g. the {\em diver} and the {\em face} sequence) and will also achieve good robustness when the size of the object does not change with respect to the initialization frame. The TTF tracker will fail uniformly, however it will produce high accuracy only when there is no rapid motion predominantly present over the entire sequence, like in sequences {\em hand}, {\em hand2}, and {\em sunshade}. The TTO tracker will achieve good robustness (no failures), however, it will not achieve good accuracy when the size of the object region changes a lot, e.g. in sequences {\em diver} and {\em gymnastics}. These observations can be extended to the entire set of sequences using clustering. As a demonstration we have used K-means clustering with expected number of clusters set to $K = 3$ to generate labels that are shown in Table~\ref{tab:properties}. The labels are of course relative to the entire set, but they summarize these relative properties well, e.g. we can see that {\em face} sequence is similar to {\em diver} sequence in terms of movement, however, the {\em diver} sequence contains a lot of size changes. This simple approach could be in future extended to provide automated and less-biased sequence descriptions.

\begin{table}[!t]
\centering
    \caption{Sequence properties according to theoretical tracker performance.}
    \label{tab:properties}
\scriptsize{
\begin{tabularx}{\columnwidth}{|l|X|X|X|X|}
\hline
&\textbf{Size \newline (TTA)}&\textbf{Motion \newline (TTS)}&\textbf{Speed \newline (TTF)}&\textbf{Size change \newline (TTO)}\\\hline
\textbf{bicycle}&small&high&medium&medium\\\hline
\textbf{biker}&large&medium&low&high\\\hline
\textbf{bolt}&small&high&medium&medium\\\hline
\textbf{can}&medium&high&medium&low\\\hline
\textbf{car}&small&medium&medium&low\\\hline
\textbf{child}&large&medium&medium&high\\\hline
\textbf{david\_indoor}&small&low&medium&low\\\hline
\textbf{david\_outdoor}&small&high&medium&low\\\hline
\textbf{dinosaur}&large&medium&low&medium\\\hline
\textbf{diver}&small&low&medium&high\\\hline
\textbf{face}&medium&low&low&low\\\hline
\textbf{gymnastics}&medium&low&medium&high\\\hline
\textbf{gymnastics2}&small&low&low&medium\\\hline
\textbf{hand}&small&high&high&medium\\\hline
\textbf{hand2}&small&high&high&medium\\\hline
\textbf{motocross1}&medium&high&medium&high\\\hline
\textbf{mountainbike}&small&medium&low&medium\\\hline
\textbf{pets2000}&small&medium&low&medium\\\hline
\textbf{pets2001-1}&small&medium&low&high\\\hline
\textbf{pets2001-2}&small&medium&low&high\\\hline
\textbf{sunshade}&small&high&high&low\\\hline
\textbf{torus}&small&high&medium&low\\\hline
\textbf{trellis}&small&low&medium&high\\\hline
\textbf{turtlebot1}&medium&medium&low&medium\\\hline
\textbf{woman}&small&medium&medium&medium\\\hline
\end{tabularx}
}
\end{table}

\section{Conclusion}
\label{conclusion}

In this paper we have addressed the problem of performance evaluation in monocular single-target short-term visual tracking. Through theoretical and experimental analysis we have investigated various popular performance evaluation measures, discussed their pitfalls and showed that many of the widely used measures are equivalent. Since some measures reflect certain aspect of tracking performance, combining those that address the same aspect provides no additional information regarding the performance or even introduces bias toward a certain aspect of performance to the result. Based on the results of our experiment we have proposed to use a pair of two existing complementary measures. This pair, that we call the A-R pair, takes into account the accuracy (using {\em average overlap}) and the robustness (using {\em failure rate}) of each tracker. We have also proposed an intuitive way of visualizing the results in a 2-dimensional scatter plot, called the A-R plot. Additionally, we have introduced fragmentation as an additional indicator for distribution of failures. We have introduced several theoretical trackers that can be used to quickly review the results of the evaluated trackers in terms of basic properties that the theoretical trackers exhibit. We have also shown that the theoretical trackers can be used for automatic annotation of sequence properties from a tracker viewpoint.

While narrowing down the abundance of performance measures is a big step toward homogenizing the tracking evaluation methodology, this is only one of the requirements for a consistent evaluation methodology for visual trackers. The measures that were proposed in this paper have already been adopted as the foundation of the evaluation methodology of a recently organized visual tracking challenges VOT2013~\cite{Kristan2013} and VOT2014~\cite{Kristan2014}, where a rigorous analysis in terms of accuracy and robustness has provided multiple interesting insights into performance of individual trackers, e.g. we have shown that some trackers are more robust, but less accurate, while some sacrifice robustness for greater accuracy. In our future work we will extend the automatic labeling of sequences using both theoretical and practically applicable trackers as well as investigate the question how to reduce the number of annotated frames without degrading the performance estimates~\cite{Carvalho2012}. 

%\vspace{0.5em}

\appendices
\section{Reformulation of CoTPS~\cite{Nawaz2012} measure \label{ap:cotps}}

Let $\phi_1, \phi_2, \dots, \phi_N$ be frame overlaps for a sequence of length $N$. In~\cite{Nawaz2012}, the CoTPS measure is defined as a weighted average of two factors, that the authors define as {\em tracking accuracy}, $\Omega$, and {\em tracking failure}, $\lambda_0$, that are combined using a dynamically computed factor, $\beta$, as

\begin{equation}
\label{eq:cotps2}
CoTPS = \beta \Omega + (1 - \beta) \lambda_0.
\end{equation}

The tracking failure factor $\lambda_0$ is computed as the percentage of frames where the tracker failed, i.e. $\lambda_0 = \frac{N_0}{N}$, where $N_0$ is a number of frames where the overlap between ground-truth region and the predicted region is $0$. The weight factor is defined as $\beta = \frac{\hat{N}}{N}$, where $\hat{N}$ denotes the number of frames where the overlap is higher than $0$, therefore $\beta = 1 - \lambda_0$. The definition for tracking accuracy part $\Omega$ is

\begin{equation}
\label{eq:cotps2-omega}
\Omega = \sum_{\tau \in (0, 1]} \frac{\hat{N}(\tau)}{\hat{N}}, 
\end{equation}

\noindent where $N(\tau) = |\{j : \phi_j \geq 0 \land \phi_j \leq \tau\}|$ denotes the number of frames that is higher than $0$, but lower than $\tau$. We observe that (\ref{eq:cotps2-omega}) is actually an approximation of the integral with respect to threshold $\tau$, that can also be reformulated as 

\begin{equation}
\label{eq:cotps2-omega2}
\Omega = \int_0^1 \frac{\hat{N}_\tau}{\hat{N}} d\tau = 1 - \int_0^1 \frac{\hat{P}_\tau}{\hat{N}} d\tau, 
\end{equation}

\noindent where $P(\tau) = |\{j : \phi_j \geq \tau \}|$. According to the proof in Appendix~\ref{ap:auc}, the integral results in average overlap over a set of frames, therefore $\Omega = 1 - \hat{\phi}$, where $\hat{\phi}$ is the average overlap over $\{\phi_j : \phi_j \geq 0 \}$. Therefore, the CoTPS measure can be rewritten as 

\begin{equation}
\label{eq:cotps3}
CoTPS = (1 - \lambda_0) ( 1 - \hat{\phi}) + \lambda_0^2.
\end{equation}

Considering that average overlap over the entire sequence can be written as $\phi = (1 - \lambda_0) \hat{\phi}$, we can further derive

\begin{equation}
\label{eq:cotps4}
CoTPS = 1 - \bar{\phi} - (1 - \lambda_0) \lambda_0,
\end{equation}

\noindent meaning that the CoTPS measure is a function of average overlap as well as the percentage of frames where the overlap is $0$.

\section{Proof that AUC of~\cite{Wu2013} equals to average overlap \label{ap:auc}}

\noindent {\bf Problem:} Let $\phi_1, \phi_2, \dots, \phi_N$ be frame overlaps for a sequence of length $N$. We assume that the frame overlaps are \emph{ordered} by scale from minimal to maximal value and $\phi_0 = 0$, i.e. 

$$ 0 = \phi_0 \leq \phi_1 \leq \cdots \leq \phi_N.$$

Let $P(\tau) = |\{j : \phi_j \geq \tau\}|$ be the number of overlaps greater than $\tau$. The AUC measure is an integral of $\frac{P(\tau)}{N}$ from $0$ to $1$. We want to prove that the average overlap, $\bar{\phi}$, for the sequence $\phi_1, \phi_2, \dots, \phi_N$ equals to the computed AUC, i.e.

$$ \frac{1}{N} \sum_{i=1}^{N} \phi_i =  \frac{1}{N} \int_0^1 P(\tau) d\tau.$$

\medskip
\medskip

\noindent {\bf Proof:} Function $P$ is a step function (constant between $\phi_i$ and $\phi_{i+1}$). Therefore its integral $I$ is

$$ I = \sum_{i=0}^{N-1} P(\phi_i) (\phi_{i+1} - \phi_{i}). $$

\noindent The sum can be reorganized in the following way:
\begin{align}
  I &= P(\phi_0)(\phi_1{-}\phi_0){+}P(\phi_1)(\phi_2{-}\phi_1){+}P(\phi_2)(\phi_3{-}\phi_2) + \dots \notag \\
    &= \phi_1 P(\phi_0){-}\phi_0 P(\phi_0){+}\phi_2 P(\phi_2){-}\phi_1 P(\phi_1){+}\phi_3 P(\phi_3){-}\dots \notag \\
    &= {-}\phi_0 P(\phi_0){+}\phi_1 (P(\phi_0){-}P(\phi_1)){+}\phi_2(P(\phi_1){-}P(\phi_2))\cdots \notag \\
    &= 0\cdot P(\phi_0) + \phi_1 \cdot 1 + \phi_2 \cdot 1 + \cdots  \label{eq:1}\\
    &= \phi_0 + \phi_1 + \phi_2 + \cdots \label{eq:2} \\
    &= \sum_{i=1}^{N} \phi_i. \notag
\end{align}

In~\eqref{eq:1} we have assumed that the shift between the two consequential values of $P(\tau)$, i.e. $P(\phi_i) - P(\phi_{i+1})$ equals to $1$, that is true if all $\phi_i$ are different. If $k$ consequential $\phi_i$ are equal then the corresponding $k-1$ shifts are $0$, while the last one is $k$. However, in~\eqref{eq:1} we add $(\phi_i\cdot 1)$ $k$ times. $\blacksquare$

%\section*{Acknowledgments}
%\noindent This research was in part supported by: ARRS projects J2-3607, J2-2221 and J2-4284.

% Can use something like this to put references on a page
% by themselves when using endfloat and the captionsoff option.
\ifCLASSOPTIONcaptionsoff
  \newpage
\fi

\bibliographystyle{IEEEtran}
\bibliography{cehovin2015tip}

% biography section
% 
% If you have an EPS/PDF photo (graphicx package needed) extra braces are
% needed around the contents of the optional argument to biography to prevent
% the LaTeX parser from getting confused when it sees the complicated
% \includegraphics command within an optional argument. (You could create
% your own custom macro containing the \includegraphics command to make things
% simpler here.)
%\begin{IEEEbiography}[{\includegraphics[width=1in,height=1.25in,clip,keepaspectratio]{mshell}}]{Michael Shell}
% or if you just want to reserve a space for a photo:

%\begin{IEEEbiography}{Michael Shell}
%Biography text here.
%\end{IEEEbiography}

% if you will not have a photo at all:
%\begin{IEEEbiographynophoto}{John Doe}
%Biography text here.
%\end{IEEEbiographynophoto}

% insert where needed to balance the two columns on the last page with
% biographies
%\newpage

%\begin{IEEEbiographynophoto}{Jane Doe}
%Biography text here.
%\end{IEEEbiographynophoto}

\vspace{-1cm}
\begin{IEEEbiography}[{\includegraphics[width=1in,height=1.25in,clip,keepaspectratio]{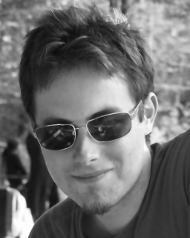}}]{Luka {\v C}ehovin} received his Ph.D from the Faculty of Computer and Information Science, University of Ljubljana, Slovenia in 2015. Currently he is working at the Visual Cognitive Systems Laboratory, Faculty of Computer and Information Science, University of Ljubljana, Slovenia as a teaching assistant and a researcher. His research interests include computer vision, HCI, distributed intelligence and web-mobile technologies.
\end{IEEEbiography}\vspace{-1cm}

\begin{IEEEbiography}[{\includegraphics[width=1in,height=1.25in,clip,keepaspectratio]{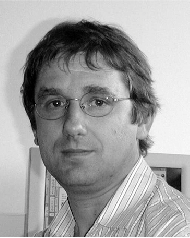}}]{Ale{\v s} Leonardis}
is a Professor at the School of Computer Science, University of Birmingham and co-Director of the Centre for Computational Neuroscience
and Cognitive Robotics. He is also a Professor at the FCIS, University of Ljubljana and adjunct professor at the FCS, TU-Graz. His research interests include robust and adaptive methods for computer vision, object and scene recognition and categorization, statistical visual learning, 3D object modeling, and biologically motivated vision.
\end{IEEEbiography} \vspace{-1cm}

\begin{IEEEbiography}[{\includegraphics[width=1in,height=1.25in,clip,keepaspectratio]{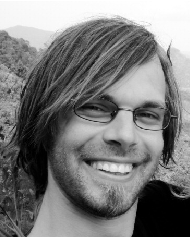}}]{Matej Kristan}
received a Ph.D from the Faculty of Electrical Engineering, University of Ljubljana in 2008. He is an Assistant Professor at the ViCoS Laboratory at the Faculty of Computer and Information Science and at the Faculty of Electrical Engineering, University of Ljubljana. His research interests include probabilistic methods for computer vision with focus on visual tracking, dynamic models, online learning, object detection and vision for mobile robotics.
\end{IEEEbiography} 

% You can push biographies down or up by placing
% a \vfill before or after them. The appropriate
% use of \vfill depends on what kind of text is
% on the last page and whether or not the columns
% are being equalized.

%\vfill

% Can be used to pull up biographies so that the bottom of the last one
% is flush with the other column.
%\enlargethispage{-5in}

% that's all folks
\end{document}